\documentclass[conference]{IEEEtran}
\usepackage{times}

\usepackage[numbers]{natbib}
\usepackage{amsmath}
\usepackage{multicol}
\usepackage[bookmarks=true]{hyperref}
\usepackage[capitalise]{cleveref}
\usepackage{xspace}
\usepackage[dvipsnames]{xcolor}
\usepackage{acronym}
\usepackage{booktabs}
\usepackage{capt-of}
\usepackage{subcaption}
\usepackage{amsfonts}
\usepackage{amssymb}
\usepackage{bm}
\usepackage{multirow}
\usepackage{graphicx} \graphicspath{{paper/figures/}}
\usepackage{mathtools}
\usepackage{algorithm}
\usepackage{algpseudocode}
\usepackage{tabularx} 
\usepackage{xcolor}
\usepackage{dblfloatfix}
\usepackage{cuted}
\usepackage{caption}
\let\labelindent\relax
\usepackage{enumitem}
\usepackage{ragged2e}
\usepackage{wasysym}
\usepackage{placeins}
\usepackage{svg}

\def\model{\textsc{OmniXtreme}\xspace}
\def\dataset{XtremeMotion\xspace}

\makeatletter
\DeclareRobustCommand\onedot{\futurelet\@let@token\@onedot}
\def\@onedot{\ifx\@let@token.\else.\null\fi\xspace}
\def\eg{\emph{e.g}\onedot}

\def\etc{\emph{etc}\onedot}

\makeatother

\acrodef{ode}[ODE]{Ordinary Differential Equation}
\acrodef{dagger}[DAgger]{Dataset Aggregation}
\acrodef{lafan1}[LAFAN1]{ Unitree-retargeted LAFAN1}
\acrodef{ppo}[PPO]{Proximal Policy Optimization}
\acrodef{fk}[FK]{Forward Kinematics}

\crefname{algorithm}{Alg.}{Algs.}
\Crefname{algocf}{Algorithm}{Algorithms}
\crefname{section}{Sec.}{Secs.}
\Crefname{section}{Section}{Sections}
\crefname{table}{Tab.}{Tabs.}
\Crefname{table}{Table}{Tables}
\crefname{figure}{Fig.}{Fig.}
\Crefname{figure}{Figure}{Figure}

\definecolor{goodgreen}{RGB}{0,150,0}
\definecolor{badred}{RGB}{180,0,0}
\definecolor{midorange}{RGB}{220,140,0}
\newcommand{\psicon}{\textcolor{badred}{\small$\circleddash$}}

\definecolor{softred}{RGB}{200, 50, 50}   % Softer red

\begin{document}

\title{OmniXtreme: Breaking the Generality Barrier in High-Dynamic Humanoid Control} %Motion Fidelity at Scale: Towards Human-Level Motion Tracking for Humanoids via Diffusion Policy

% You will get a Paper-ID when submitting a pdf file to the conference system
\author{
\authorblockN{
Yunshen Wang$^{1,2,3,*}$,
Shaohang Zhu$^{1,2,4,*}$,
Peiyuan Zhi$^{1,2}$,
Yuhan Li$^{1,2,6}$,
Jiaxin Li$^{1,2,7}$,\\
Yong-Lu Li$^{3}$,
Yuchen Xiao$^{5}$,
Xingxing Wang$^{5}$,
Baoxiong Jia$^{1,2,\dagger}$,
Siyuan Huang$^{1,2,\dagger}$
}

\authorblockA{$^{1}$State Key Laboratory of General Artificial Intelligence, Beijing Institute for General Artificial Intelligence (BIGAI)}
\authorblockA{$^{2}$Joint Laboratory of Embodied AI and Humanoid Robots, BIGAI \& Unitree Robotics}

\authorblockA{
$^{3}$Shanghai Jiao Tong University \quad
$^{4}$University of Science and Technology of China
}

\authorblockA{
$^{5}$Unitree Robotics \quad
$^{6}$Huazhong University of Science and Technology \quad
$^{7}$Beijing Institute of Technology
}
\authorblockA{
$^{*}$Equal contribution. \quad
$^{\dagger}$Corresponding authors.
}
\authorblockA{
Project page: \url{https://extreme-humanoid.github.io/}
}
}

%\author{\authorblockN{Michael Shell}
%\authorblockA{School of Electrical and\\Computer Engineering\\
%Georgia Institute of Technology\\
%Atlanta, Georgia 30332--0250\\
%Email: mshell@ece.gatech.edu}
%\and
%\authorblockA{Twentieth Century Fox\\
%\and
%\authorblockN{James Kirk\\ and Montgomery Scott}
%\authorblockA{Starfleet Academy\\
%San Francisco, California 96678-2391\\

% avoiding spaces at the end of the author lines is not a problem with
% conference papers because we don't use \thanks or \IEEEmembership

%Homer Simpson\authorrefmark{2},
%Montgomery Scott\authorrefmark{3} and
%Eldon Tyrell\authorrefmark{4}}
%Telephone: (800) 555--1212, Fax: (888) 555--1212}
%\authorblockA{\authorrefmark{4}Tyrell Inc., 123 Replicant Street, Los Angeles, California 90210--4321}}

\maketitle

\begin{strip}
\centering
\begin{minipage}{.95\textwidth}
  \centering
  \includegraphics[width=\linewidth]{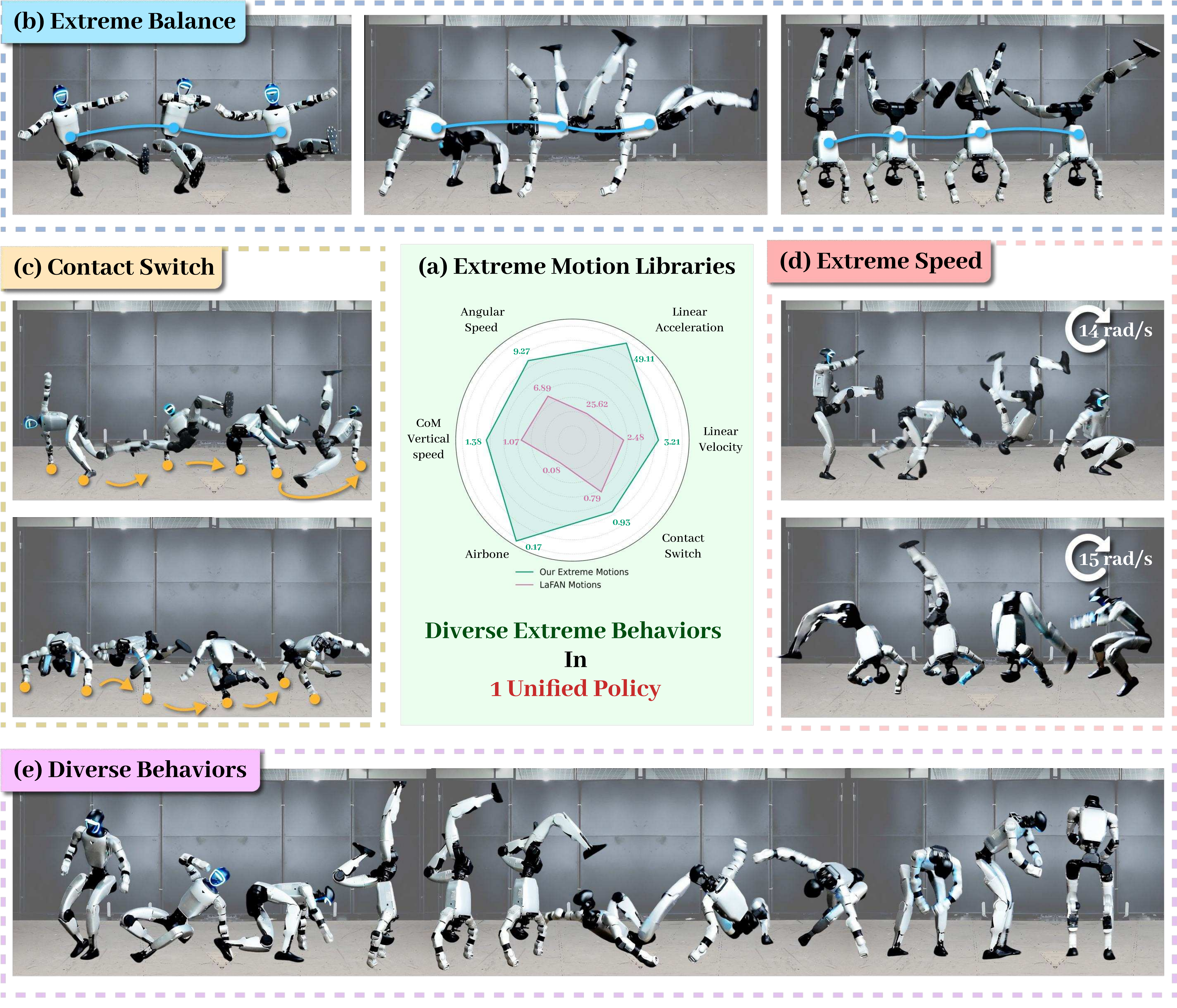}
  \captionof{figure}{\small \textbf{Extreme whole-body humanoid control from our unified policy \model.}
  (a) A quantitative comparison shows that our curated extreme motion libraries occupy
  substantially more challenging regimes than standard multi-motion benchmarks
  (\eg, Unitree-retargeted LAFAN1). Real-world executions of our unified policy \model demonstrate robust and physically executable extreme behaviors drawn from this motion set,
  including (b) extreme balance behaviors,
  (c) rapid contact switching with complex support transitions,
  (d) high-speed motions with large angular velocities,
  and (e) diverse whole-body behaviors spanning qualitatively distinct motion styles.}
  \label{fig:figure1}
\end{minipage}
\end{strip}
\begin{abstract}

High-fidelity motion tracking serves as the ultimate litmus test for generalizable, human-level motor skills. However, current policies often hit a ``generality barrier'': as motion libraries scale in diversity, tracking fidelity inevitably collapses—especially for real-world deployment of high-dynamic motions. We identify this failure as the result of two compounding factors: the learning bottleneck in scaling multi-motion optimization and the physical executability constraints that arise in real-world actuation. To overcome these, we introduce \model, a scalable framework that decouples general motor skill learning from sim-to-real physical skill refinement. Our approach uses a flow-matching policy with high-capacity architectures to scale representation capacity without the interference-intensive multi-motion RL optimization, followed by an actuation-aware refinement phase that ensures robust performance on physical hardware. Extensive experiments demonstrate that \model maintains high-fidelity tracking across diverse, high-difficulty datasets. On real robots, the unified policy successfully executes multiple extreme motions, effectively breaking the long-standing fidelity–scalability trade-off in high-dynamic humanoid control.

\end{abstract}
\IEEEpeerreviewmaketitle

\section{Introduction}
We ultimately seek general-purpose humanoids with scalable, human-level whole-body motor skills. A natural and widely used way to study such capability is high-fidelity motion tracking, where a controller must reproduce reference motions accurately while remaining dynamically stable under contacts and disturbances. High-quality tracking is more than an aesthetic goal: it captures whole-body coordination and contact timing that underlie loco-manipulation, expressive interaction, and many downstream core humanoid capabilities~\cite{zhao2025resmimic,yang2025omniretarget,yin2025visualmimic,allshire2025visual,su2025hitter,weng2025hdmi,yu2025skillmimic}.

Over the past years, learning-based motion tracking has made striking progress: with carefully designed objectives and reinforcement learning, policies can track individual motions with high precision, including highly dynamic behaviors such as dance, flips, and martial arts~\cite{zhang2025hub,liao2025beyondmimic,he2025asap}. More recent work~\cite{chen2025gmt,luo2025sonic,fu2024humanplus,ji2024exbody2,he2024omnih2o,cheng2024expressive,he2024learning,zhang2025track,ze2025twist,li2025clone} has taken important steps toward multi-motion controllers that cover broader behavior libraries. Yet a recurring pattern persists: when we scale to larger, more heterogeneous motion libraries spanning diverse styles, contact regimes, and timing modes, motion tracking quality tends to degrade. Controllers become conservative and ``average,'' break on the hardest motions, or prove brittle to the small deviations that inevitably occur in sim-to-real transfer. The degradation is particularly pronounced in \emph{high-dynamic motions},
where even small tracking errors can rapidly cascade into catastrophic failures.
This long-standing fidelity--scalability trade-off has effectively capped the
level of generality achievable in humanoid motor control, particularly in
\emph{high-dynamic regimes}, suggesting a fundamental limitation rather than
isolated engineering issues~\cite{he2024omnih2o,zhang2025track,he2025asap,zhang2025hub}.

A central question therefore arises: \textbf{why is high-fidelity motion tracking so difficult to scale, especially on real humanoid robots?}
We argue that this difficulty stems from two compounding barriers that emerge at different stages of the current sim-to-real training pipeline.

The first barrier is the \textbf{learning bottleneck} that arises even in simulation. Several recent works~\cite{zhang2025track,chen2025gmt,ji2024exbody2,luo2025sonic,ze2025twist,li2025clone,he2024learning,he2024omnih2o} have begun to explore multi-motion humanoid tracking, aiming to improve scalability beyond single-motion imitation. However, existing approaches remain constrained by both representation and optimization. On the representation side, most approaches rely on relatively simple policy parameterizations, such as MLP actors~\cite{zhang2025track,chen2025gmt,ji2024exbody2,ze2025twist,li2025clone,he2024learning,he2024omnih2o,wang2025experts}. When required to map observations to highly heterogeneous action targets arising from
diverse behaviors and contact patterns, such parameterizations have been observed to
exhibit limited scalability as data diversity increases~\cite{pan2025much}. On the optimization side, jointly training a unified policy across many motions using reinforcement learning exacerbates gradient interference, often leading to conservative averaging and selective failures on \emph{high-dynamic behaviors}~\cite{he2024learning,he2024omnih2o,fu2024humanplus,luo2025sonic}. Together, these factors cause tracking fidelity to collapse as motion diversity and difficulty increase.

The second barrier is the \textbf{physical executability bottleneck} that emerges at deployment. Even when high-fidelity tracking is achieved in simulation, transferring such behaviors to physical robots remains challenging. In prior humanoid learning pipelines~\cite{he2024omnih2o,zhang2025track,chen2025gmt,ji2024exbody2,luo2025sonic,ze2025twist,liao2025beyondmimic,he2025asap,li2025clone,xie2025kungfubot}, actuation constraints during training are modeled primarily through joint position limits and simple effort bounds. Although these simplifications facilitate learning, they become insufficient in \emph{high-dynamic motions}, where system behavior is dominated by unmodeled actuator nonlinearities~\cite{shin2023actuator}, such as torque--speed characteristics and velocity-dependent torque losses, as well as power-related effects, including regenerative power phenomena, leading to rapid degradation of execution stability. As a result, fidelity that appears scalable in simulation may still fail to materialize on real robots.

Motivated by this analysis, we propose \model, a scalable training framework designed to explicitly address both barriers, with the goal of enabling a single
policy to robustly control diverse and \emph{high-dynamic} humanoid behaviors. To overcome the learning bottleneck, \model adopts a flow matching policy and performs specialist-to-unified generative pretraining via behavior cloning from a collection of motion specialists. This design decouples representation learning from optimization, scaling expressive capacity through a high-capacity generative policy while avoiding interference-heavy multi-motion reinforcement learning.

To overcome the physical executability bottleneck, \model introduces a residual reinforcement learning post-training refinement for execution under realistic actuation constraints, which become particularly critical in
\emph{high-dynamic motions}. Rather than relearning motion tracking, this stage refines the pretrained policy to respect real-world actuation constraints through actuation-aware modeling, refined domain randomization, and explicit penalties on power-related effects. This targeted refinement ensures that the scaled tracking policy remains physically executable under realistic hardware dynamics.

We validate \model through extensive simulation and real-robot evaluations on
increasingly diverse and high-dynamic motion libraries.
Beyond standard multi-motion benchmarks, we curate a set of extreme motions
characterized by high speed, frequent contact transitions, and tight timing constraints,
and evaluate \model across this full spectrum.
As shown in~\cref{fig:figure1}, \model successfully executes a wide range of
extreme behaviors on a Unitree G1 humanoid robot, including flips, acrobatics,
and breakdancing where even minor deviations can rapidly cascade into failure.
Together, these results constitute a stringent scalability stress test and challenge the prevailing assumption that tracking fidelity must collapse as motion diversity and difficulty increase.

% Overall, our contributions can be summarized as follows:
% \begin{enumerate}
% \item We identify and analyze two compounding barriers that limit the scalability of high-fidelity humanoid motion tracking on real robots: a learning bottleneck arising from representation and optimization in multi-motion training, and a physical executability bottleneck that emerges at deployment.
% \item We propose a scalable motion tracking pipeline based on specialist-to-unified generative pretraining, which decouples representation capacity from optimization and enables a single policy to track diverse, high-difficulty motions without fidelity collapse.
% \item We introduce a deployment-oriented residual reinforcement learning post-training stage that incorporates actuator-aware constraints and refined domain randomization, ensuring that the scaled tracking policy remains physically executable under real-world dynamics.
% \item We demonstrate extensive results in simulation and on real hardware, including the execution of multiple high-difficulty motions by a single unified policy, providing a stringent stress test that challenges the prevailing fidelity--scalability trade-off.
% \end{enumerate}
Overall, our contributions are fourfold:
\begin{enumerate}[leftmargin=*,noitemsep,nolistsep,topsep=0pt,partopsep=0pt]
\item We present \model, a scalable training framework for high-fidelity humanoid
motion tracking that explicitly tackles the fundamental scalability challenge in
\emph{high-dynamic} humanoid control.
\item We introduce a specialist-to-unified generative pretraining stage based on flow matching, enabling a unified policy to scale across heterogeneous and high-dynamic motions.
\item We propose an actuation-aware residual reinforcement learning post-training stage that refines the pretrained policy under realistic actuation constraints, ensuring physical executability.
\item We demonstrate through extensive simulation and real-world experiments that \model enables a single unified policy to robustly execute diverse and extreme motions, addressing the conventional fidelity--scalability trade-off, especially for high-dynamic motions.
\end{enumerate}

\section{Related Work}

% general humnoid motion tracker  /  humanoid whole body control
%  certain skill， human getup， humanoid parkour
% imitate sequence beyond，asap  certain sequence ，not scalable in tons of dat
%  omnih2o，exbody2 ，gmt ，any2track， scalable，but mlp based，  fidelity–scalability trade-off 
% 

% generative（diffusion based） action modeling for robot control/Diffusion（generative） Models for Robotic Planning and Control

 % DiffuseLoco takes a step in this direction by applying diffusion models to multi-skill learning on quadruped robots, the range of skills demonstrated is still limited, falling short of fully showcasing the expressive capacity of diffusion-based policies
 
% beyondmimc，diffusion， howerever， it focus more on guidance control，and work less on  scalablity and agility
% 

% agile robotics control

% however seldom  work on humanoid control

% residual learning in robotic lowlevel control

% Arm based policy，Policy Decorator and ResiP use  reisdual policy for 
% resmimic、 for down stream task， but not for a better policy performance
% our refinement and adaption for realworld deployment for humanoid

\subsection{Humanoid Whole-body Control and General Tracking}
Recent research in humanoid whole-body control has demonstrated remarkable progress across diverse skills~\cite{zhuang2024humanoid,huang2025learning,humanup25,zhang2025hub,overchall}, including dance, fall recovery, and parkour. 
However, achieving both high-fidelity motion tracking and scalability across large and diverse motion libraries remains an open challenge. 
Frameworks such as ASAP~\cite{he2025asap} and BeyondMimic~\cite{liao2025beyondmimic} demonstrate strong performance in high-quality imitation of individual motion clips, yet extending these approaches to increasingly large motion sets introduces additional optimization complexity. 
On the other hand, large-scale RL-based trackers including OmniH2O~\cite{he2024omnih2o}, ExBody2~\cite{ji2024exbody2}, and GMT~\cite{chen2025gmt} show promising scalability, though maintaining precise motion fidelity under extensive skill coverage remains challenging. 
This tension is often reflected as a fidelity–scalability trade-off in practice. 
To address this issue, \model introduces a generative action representation and a specialist-to-unified optimization framework, enabling scalable learning while maintaining strong tracking precision across high-dynamic motion datasets.

\subsection{Diffusion and Flow-based Action Modeling for Robotic Planning and Control}

Diffusion and flow-based models~\cite{song2020denoising, ho2020denoising, peebles2023scalable,rombach2022high,lipman2022flow,pan2025much,tessler2025maskedmanipulator,tessler2024maskedmimic,chi2023diffusion} have shown strong capability in robot learning, leveraging iterative refinement and stochastic sampling to enhance robustness and diversity in robotic control and planning~\cite{pan2025much}. While early research focused on high-level trajectory planning or low-frequency visuomotor tasks~\cite{janner2022planning,huang2023diffusion,chi2023diffusion,xian2023chaineddiffuser,black2023zero}, DiffuseLoco~\cite{huang2024diffuseloco} takes a step to apply them to high-frequency quadruped control. To further enhance expressivity and robustness, recent works like Policy Decorator~\cite{yuan2024policy} and ResiP~\cite{ankile2025imitation} introduce residual policy learning on arm-based robots, coupling frozen base models with refinement layers to handle covariate shift and precision bottlenecks in long-horizon assembly. However, given the vast skill space and inherent instability that distinguish humanoids from quadrupeds and manipulators, current effort such as BeyondMimic~\cite{liao2025beyondmimic} focuses on guided control interfaces rather than the scalability and high-speed agility essential for high-dynamic humanoid motion tracking. Different from previous work, \model introduces a comprehensive training pipeline involving DAgger-based Flow Matching pretraining and residual post-training that pushes the boundaries of low-level scalability and agility, far surpassing the motion diversity and dynamic performance of previous approaches.

% \subsection{Residual Learning in Low-level Control}
% Residual learning has been effectively employed in robotic manipulation to refine pre-trained imitation models. Works like Policy Decorator\cite{yuan2024policy} and ResiP\cite{ankile2025imitation} utilize lightweight residual offsets on frozen base policies to mitigate covariate shift and precision bottlenecks in long-horizon assembly. In the humanoid domain, ResMimic\cite{zhao2025resmimic} employs residual strategies for task-specific transfer rather than intrinsic performance enhancement. ASAP\cite{he2025asap} employs residual learning to capture the dynamics gap between simulation and the real world by collecting real-world data; while this approach is principled, it is difficult to scale. In contrast, our framework utilizes a Residual Policy to fine-tune the base model’s performance in the simulator, specifically adapting to non-linear actuator dynamics and motor torque limits. This ensures agile, high-precision control that aligns simulation-trained priors with real-world physical constraints.

\subsection{Actuation-aware Agile Robotic Control}
Achieving agility remains a frontier in robotics~\cite{kim2025high,li2025reinforcement,he2024agile,kim2019highly,hwangbo2019learning,margolis2024rapid,zhang2024learning,hoeller2024anymal,cheng2024extreme,zhuang2023robot,zhuang2024humanoid,xie2025kungfubot}. ACRL~\cite{shin2023actuator} leveraged actuator-constrained RL for high-speed quadrupedal locomotion, while Closing the Reality Gap~\cite{zhao2026closing} utilized a current-to-torque calibration and actuator dynamics modeling for dexterous five-finger manipulation. Despite these advancements in other morphologies, learning agile and actuation-aware control policies for humanoids remains an underexplored area. \model addresses this gap by integrating physics-informed motor modeling and actuation regularization, pushing the boundaries of agile humanoid performance under realistic hardware constraints.
% To bridge this gap, \model incorporates physics-informed motor modeling and actuation regularization, thereby pushing the boundaries of agile humanoid performance under realistic hardware constraints.
%  if it 
% pretrain and post-train
\begin{figure}[!t]
\centering
\includegraphics[width=.995\columnwidth]{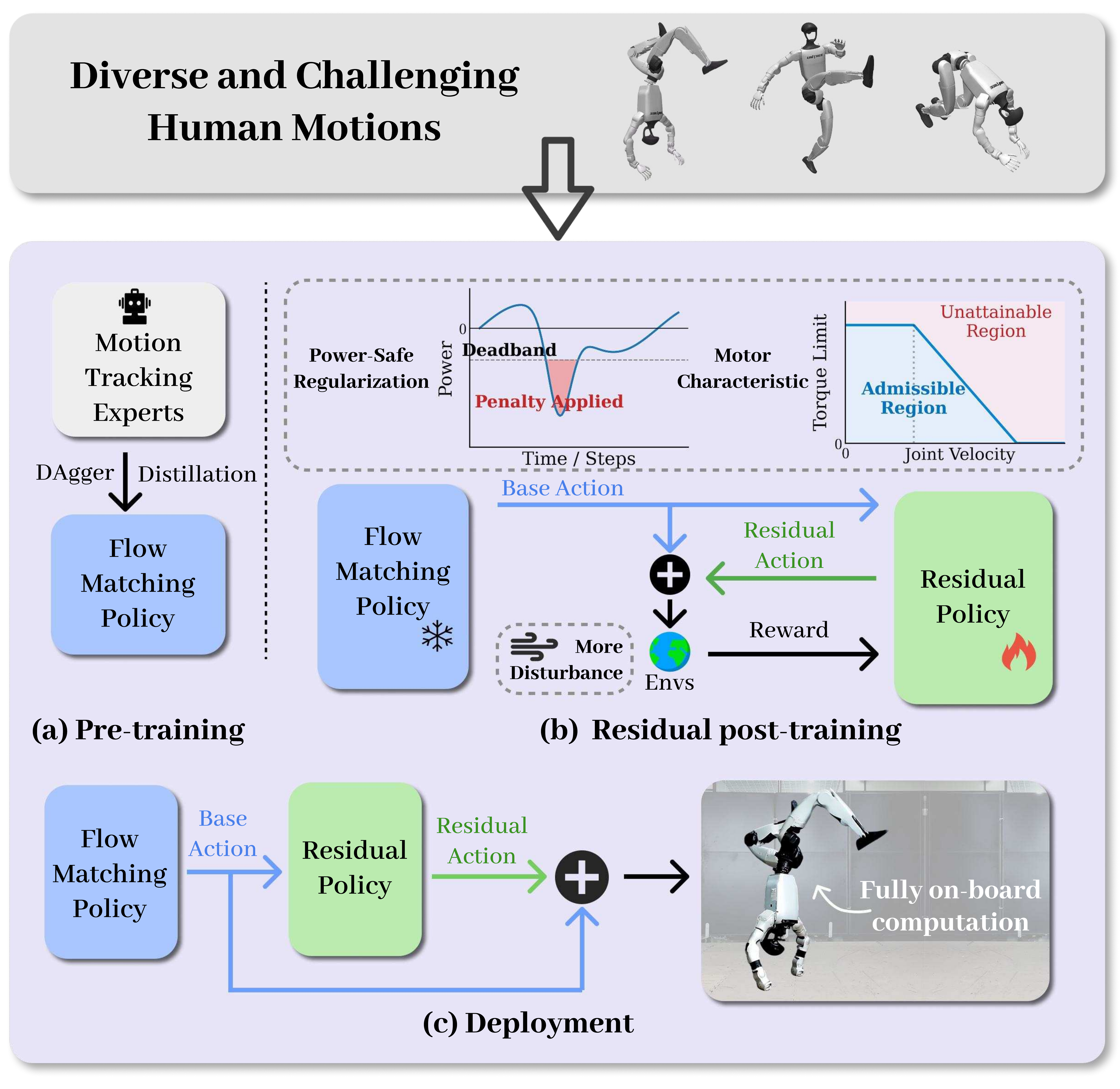}
\caption{\small \textbf{Overview of the \model.} (a) Pretraining phase: A unified base policy is trained via DAgger-based Flow Matching to aggregate diverse motion priors from different motion tracking experts. (b) Post-training phase: The base policy is frozen while a residual policy is optimized under stringent motor constraints, extensive domain randomization, and power-safety regularization to bridge the sim-to-real gap. (c) Onboard deployment: The whole inference pipeline is real-time and executed entirely onboard, facilitating robust and agile control in physical environments.}
\label{fig:method_overview}
\vspace{-17pt}
\end{figure}

\section{Methodology}
% (to refine, stress key point of the article; to refine, repeat main ideas while explain methods;to refine. the logic between each parts; and symbols, align with graph, maybe the order of each part need changes as well )\\
In this section, we present \model, a two-stage training framework for scalable, high-fidelity humanoid motor skill learning. 
The \textbf{Scalable Flow-based Pretraining} stage focuses on high-fidelity motion imitation and representation capacity acquisition. 
Specifically, we distill diverse expert behaviors from a collection of motion-specific expert policies into a single unified base policy using flow matching~\cite{lipman2022flow}. 
This generative pretraining stage establishes a shared tracking prior across heterogeneous motions, without relying on interference-prone joint multi-motion reinforcement learning.

To address the gap between simulation and real-world execution, we further introduce an \textbf{Actuation-Aware Post-Training} stage based on residual reinforcement learning. 
Rather than relearning motion tracking, a residual policy is trained to produce corrective actions that complement the pretrained flow matching base policy. This stage aligns the overall system with real-world actuation constraints while introducing substantially more aggressive domain randomization.
Through this targeted refinement, the residual policy adapts the pretrained tracking behavior to realistic hardware dynamics, improving physical executability and deployment robustness.

\subsection{Scalable Flow-based Policy Pretraining} 
 
\subsubsection{Problem Formulation} During the pretraining phase, we learn a flow-matching robot policy with \ac{dagger}-based distillation~\cite{ross2011reduction,tessler2025maskedmanipulator}. Specifically, we consider the observation space of $o = \{p, c, h\}$ covering: (i) robot proprioception $p$, including joint positions, velocities, base angular velocities, and previous actions; (ii) command $c$ consisting of 6D torso orientation differences along with target joint positions and velocities from the reference motion; and (iii) history information $h$ encompassing past proprioceptive states. Given a reference motion dataset $M = \{m_i\}_{i=1}^{M}$, our goal is to first learn expert policies $\Pi_{\text{expert}}^i = \{\pi_{\text{expert}}^i(a | o)\}_{i=1}^{M}$ for each reference motion, then distilling it into a flow-based general policy $\pi_\theta(a|o)$.

\subsubsection{Expert Policy Learning} For expert policy training, we draw the reference motion dataset $M$ from a combination of \ac{lafan1} dataset~\cite{harvey2020robust}, AMASS~\cite{mahmood2019amass}, MimicKit~\cite{peng2025mimickit}, and the Reallusion motion library~\cite{reallusion}, covering both diverse behavioral patterns and high-dynamic maneuvers. All reference motions are first retargeted to the Unitree G1 humanoid robot using GMR~\cite{ze2025twist,joao2025gmr}. Subsequently, we train each expert policy $\pi_{\text{expert}}^k$ on the specific motion $m_k$ via \ac{ppo}~\cite{schulman2017proximal}.

% The objective of pre-training is to establish a high-performance base policy capable of diverse motions. We first train a collection of motion-specific expert policies via \ac{ppo}\cite{schulman2017proximal}, each tracking a single motion or a small subset of closely related motions. We then apply flow matching to distill these experts into a unified conditional action distribution. This pretraining stage prioritizes high-fidelity motion imitation, establishing a strong representational foundation before subsequent residual reinforcement learning refinement.

% \noindent\textbf{Motion Processing and Teacher Training}
% Our motion dataset is drawn from a combination of the \ac{lafan1} dataset\cite{harvey2020robust}, AMASS\cite{mahmood2019amass}, MimicKit\cite{MimicKitPeng2025}, and the Reallusion motion library\cite{reallusion}, covering both diverse behavioral patterns and high-dynamic maneuvers. All reference motions are first retargeted to the Unitree G1 humanoid robot using GMR\cite{ze2025gmr}. Subsequently, we trained a diverse set of teachers across retargeted motions, obtaining a representative expert distribution that spans a wide range of motor behaviors, which serves as the supervision source for subsequent \ac{dagger}-based distillation\cite{ross2011reduction,tessler2025maskedmanipulator}.

\subsubsection{Flow-matching Policy Learning}
We learn the flow-matching robot policy with \ac{dagger} by first rolling out the current flow-based policy $\pi_\theta(a|o)$ in the simulator and collecting a trajectory of visited states $\{o_1, \cdots, o_N\}$ given reference motion dataset $M$. For each visited state $o$, we obtain the expert action $a_{\text{expert}}$ by querying the corresponding expert policy. The flow-based model then learn to recover the expert action $a_{\text{expert}}$ from noised action by optimizing:
\begin{equation}
\label{eq:flow_matching}
\mathcal{L}_{\text{FM}}(\theta) = \mathbb{E}_{t, \epsilon, a_{\text{expert}}} \left[ \| v_\theta(a_t, t, o) - (\epsilon - a_{\text{expert}}) \|^2 \right],
\end{equation}
where $a_t = (1 - t) a_{\text{expert}} + t \epsilon$ is the noised action interpolated between expert action $a_{\text{expert}}$ and random noise $\epsilon\sim\mathcal{N}(0,I)$ depending on flow timestep $t\in[0,1]$. This objective learns a velocity field $v_\theta(a_t,t,o)$ to predict the target velocity $u = \epsilon - a_{\text{expert}}$, learning the denoising directions at each flow timestep~\cite{lipman2022flow}. During the optimization process, the timestep $t$ is sampled from a Beta distribution, $t \sim \mathrm{Beta}(\alpha, \beta)$, to focus the learning process on specific regions of the probability path, thereby enhancing convergence and trajectory refinement. With the velocity field $v_\theta$, we can generate action $a_0$ from random noise $a_1 \sim \mathcal{N}(0,I)$ by integrating $v_\theta$ from $t=1$ to $t=0$ via the foward Euler rule:
\begin{equation}
a_{t-\frac{1}{D}} = a_t - \frac{1}{D}v_\theta(a_t,t,o),
\end{equation}
where $D$ is the number of integration or denoising steps controlling the approximation accuracy.  By iteratively rolling out trajectories and supervising them with expert actions using \cref{eq:flow_matching}, we learn $\pi_\theta$ as a general policy to map the current observation to appropriate actions. The full training procedure is illustrated in~\cref{fig:method_overview}(a) and detailed in~\cref{alg:DAgger_fm}.

\subsubsection{Fidelity-Preserving Randomization and Noise}
To maintain a high degree of motion expressivity while ensuring physical stability, we implement a conservative randomization and noise strategy, as detailed in \cref{tab:randomization_params_standard_size}, during both the teacher training and pretraining phases. By utilizing moderate noise levels and domain randomization, we prevent the performance collapse often induced by excessive stochasticity. This ensures that the agent accurately captures the underlying physical dynamics, resulting in a flow matching policy that possesses foundational sim-to-real robustness and the predictive certainty necessary for real-world deployment.
% \begin{table}[htbp]
% \centering
% \caption{Configurations for Moderate Noise and Domain Randomization}
% \label{tab:randomization_noise_params}
% \begin{tabular}{@{}lc@{}}
% \toprule
% \textbf{Parameter Item} & \textbf{Value / Range} \\ \midrule
% Joint Position Noise & [Value] \\
% Joint Velocity Noise & [Value] \\
% Angular Velocity Noise & [Value] \\
% Base Linear Velocity Noise & [Value] \\ % 新增项
% Base COM Offset & [Value] \\ % 新增项
% Default Calibration Noise & [Value] \\
% Initial Pose Noise & [Value] \\
% Initial Velocity Noise & [Value] \\
% Push Frequency & [Value] \\
% Push Magnitude & [Value] \\ \bottomrule
% \end{tabular}
% \end{table}

% \input{tables/table_method_moderate_noise}
\begin{table}[t]
\centering
% \small
% \footnotesize
\scriptsize
% \resizebox{\textwidth}{!}
\caption{\small Configurations for noise, domain randomization, and termination thresholds during pre-training and post-training phases. Here $\pm x$ denotes $[-x, x]$.}
\label{tab:randomization_params_standard_size}
\renewcommand{\arraystretch}{1.2} 
\begin{tabularx}{\columnwidth}{@{} >{\RaggedRight\arraybackslash}p{3.2cm} !{\hspace{-15pt}} l !{\hspace{-11pt}} X @{}}
\toprule
\textbf{Parameter Item} & \textbf{Moderate} & \textbf{Aggressive} \\ \midrule
\multicolumn{3}{l}{\textit{Noise and Domain Randomization}} \\
Joint Position (rad) & $\pm 0.01$ & $\pm 0.01$ \\
Joint Velocity (rad/s) & $\pm 0.5$ & $\pm 0.5$ \\
Angular Velocity (rad/s) & $\pm 0.2$ & $\pm 0.2$ \\
Torso 6D Rotation (rad) & $\pm 0.05$ & $\pm 0.05$ \\
Base CoM Offset (m) & $x$: $\pm 0.025, y,z$: $\pm 0.05$ & $x$: $\pm 0.025, y,z$: $\pm 0.05$ \\
Static Friction & $[0.3, 1.6]$ & $[0.3, 1.6]$ \\
Dynamic Friction & $[0.3, 1.2]$ & $[0.3, 1.2]$ \\
Action Delay (ms) & $[0, 15]$ & $\mathbf{[5, 10]}$ \\
Coefficient of Restitution & None & $\mathbf{[0.0, 0.5]}$ \\
Default Calib. (rad) & $\pm 0.01$ & $\pm 0.01$ \\
Init. Pose (rad) & $\pm 0.1$ & $\mathbf{\pm 0.15}$ \\
Init. Lin. Vel. (m/s) & $xy$: $\pm 0.5, z$: $\pm 0.2$ & $\mathbf{xy: \pm 0.75, z: \pm 0.3}$ \\
Init. Ang. Vel. (rad/s) & $RP$: $\pm 0.52, Y$: $\pm 0.78$ & $\mathbf{RP: \pm 0.78, Y: \pm 1.17}$ \\
Push Frequency (s) & $1.0 - 3.0$ & $1.0 - 3.0$ \\
Push Lin. Vel. (m/s) & $xy$: $\pm 0.5, z$: $\pm 0.2$ & $xy$: $\pm 0.5, z$: $\pm 0.2$ \\
Push Ang. Vel. (rad/s) & $RP$: $\pm 0.52, Y$: $\pm 0.78$ & $RP$: $\pm 0.52, Y$: $\pm 0.78$ \\
Terrain Surface / Step (m) & None & $\mathbf{[0, 0.01] / 0.01}$ \\ \midrule
\multicolumn{3}{l}{\textit{Termination Thresholds}} \\
Torso Pos. Z / Ori. Error & $0.25$m / $0.8$rad & $\mathbf{0.375m / 1.2rad}$ \\
End-Effector Z-Error (m) & $0.25$ & $\mathbf{0.375}$ \\ \bottomrule
\end{tabularx}
\vspace{-17pt}
\end{table}

\begin{algorithm}[t]
\caption{Flow-based Pretraining and Inference}
\label{alg:DAgger_fm}
% \small
\begin{algorithmic}[1]

\State \textbf{Training: Distill Flow Matching Policy with \ac{dagger}}
\State \textbf{Input:} Teacher policy set $\pi_{teacher}$, Flow matching policy $\pi_{\theta}$, Motion dataset $\mathcal{M}$, Replay buffer $\mathcal{D}$
\Repeat
    \State $\mathcal{D} \leftarrow \emptyset$ \Comment{On-policy reset: Clear buffer for the new iteration}
    % \State $s_{1:T} \leftarrow \text{Rollout } \pi_{\theta} \text{ in simulator tracking } m \in \mathcal{M}$
    % \State $a_{expert} \leftarrow \pi_{teacher}(s_t, m)$ for all $t \in \{1 \dots T\}$ \Comment{Expert labeling}
    \State Sample motion $m \sim \mathcal{M}$ and select teacher $\pi^{m}_{teacher}$
    \State Rollout $\pi_{\theta}$ in simulator conditioned on $m$ to obtain states $s_{1:T}$
    \For{$t=1$ to $T$}
        \State $a_{expert,t} \leftarrow \pi^{m}_{teacher}(s_t)$ \Comment{Expert labeling}
        \State $\mathcal{D} \leftarrow \mathcal{D} \cup \{(s_t, m, a_{expert,t})\}$ \Comment{Aggregate data}
    \EndFor
    
    \State \textbf{Flow Matching Optimization:}
    \For{each gradient step}
        \State Sample $(s, a_{expert}) \sim \mathcal{D}$
        \State Sample $t \sim \text{Beta}(\alpha, \beta)$ and $\epsilon \sim \mathcal{N}(0, I)$
        \State Construct probability path: $x_t = (1-t)a_{expert} + t\epsilon$
        \State Compute target velocity: $u_t = \epsilon - a_{expert}$
        \State Update $\theta$ using gradient descent on $\| v_{\theta}(x_t, t, c) - u_t \|^2$
    \EndFor
\Until{convergence}

\Statex
\State \textbf{Inference: Action Sampling with Euler Integration}
\State \textbf{Input:} Trained velocity field $v_{\theta}$, Concatenated condition $c$, Number of steps $N$
\State Set step size $\Delta t = 1/N$
\State Initialize $x_1 \sim \mathcal{N}(0, I)$ \Comment{Start from Gaussian noise}
\For{$k = 0$ to $N-1$}
    \State $t = 1 - k \cdot \Delta t$ \Comment{Reverse time from 1 to 0}
    \State $v_t \leftarrow v_{\theta}(x_t, t, c)$
    \State $x_{t-\Delta t} \leftarrow x_t - v_t \cdot \Delta t$ \Comment{Reverse-time Euler step to obtain $x_0$}
\EndFor
\State \Return $a = x_0$ \Comment{Execute final reconstructed action}
\end{algorithmic}
\end{algorithm} 
\subsection{Actuation-Aware Post-training Phase}
% Advantages of Residual Refinement:
% 1.	Stability: The base policy provides a stable "foundation" of expert behavior. $a_{res}$ only provides small-scale corrections, ensuring the robot's fundamental logic remains intact and movements remain graceful. \textbf{RL more stable since prior}
% 2.	Efficiency: The MLP’s small parameter footprint allows RL to converge rapidly for high-precision task adaptation.
% 3.	Preservation of Diversity: Traditional RL fine-tuning often collapses a multimodal distribution into a single optimal solution. By using $a = a_{base} + a_{res}$, the residual policy fine-tunes the specific "peak" of the current distribution without destroying the rich behavioral diversity stored in the base model. \textbf{Multimodal remains sinced this design}

 \subsubsection{Residual Policy Modeling} While the pretrained flow matching base policy provides a robust and unified behavioral foundation, it encounters performance gaps when facing real-world physics. To better account for this gap and enable smooth sim-to-real transfer, we propose a post-training refinement stage using a lightweight MLP-based residual-corrective learning. Specifically, we learn the residual correction policy $\pi_\phi$ on top of the frozen pretrained policy $\pi_\theta$ by generating the refined action $a = a_{\text{flow}} + a_{\text{res}}$ and supervising it with cumulative rewards via PPO, detailed in the Appendix. 

In particular, the observation space for the residual actor and critic incorporates robot proprioception, motion command, and the current base action $a_\text{flow}$. Within the proprioceptive state, the residual policy observes the previous refined action, whereas the flow matching base policy remains conditioned on the previous flow-based action.

 \subsubsection{Actuation-aware Physical Constraint Modeling}
To explicitly account for real-world actuation effects, we train the residual policy
using environments that incorporate realistic actuation-aware physical constraints
and domain randomization, as shown in \cref{fig:method_overview}(b). The actuation-aware physical modeling is detailed as follows:
% \noindent\textbf{Domain Adaptation on Physical Constraints}
% To bridge the sim-to-real gap during residual policy refinement, we adopt a structured refinement strategy that jointly expands the disturbance envelope, discourages hardware-adverse behaviors, and enforces realistic actuator constraints. Compared to the base policy pre-training stage, this phase is intentionally more challenging, exposing the controller to conditions that better approximate unrefined real-world physics.

\paragraph{Aggressive Domain Randomization}
We substantially increase the range for domain randomization by up to 50\% on common domain randomization settings, including initial pose noise, force disturbances magnitude, angular velocity, \etc, as detailed in~\cref{tab:randomization_params_standard_size}. We randomize the terrain by adding surface noise and placing vertical steps randomly in the scene. Crucially, we relax the termination thresholds by 1.5$\times$ from their base values (\eg, orientation error from 0.8 to 1.2\,rad). This relaxation allows the residual policy to explore and correct for large-deviation but recoverable states that would otherwise be prematurely terminated.% rather than overly conservative avoidance strategies.

\paragraph{Power-Safe Actuation Regularization}
In practice, highly dynamic motions can induce large transient braking loads that are not explicitly regulated in standard training pipelines. To address the issue, we introduce an explicit penalty on excessive negative joint mechanical power to mitigate aggressive motor braking that can trigger overcurrent protection or thermal stress on real robots. Specifically, we use the instantaneous mechanical power $P = \tau\cdot\omega$ calculated from the applied joint torque $\tau$ and angular velocity $\omega$ as a critical policy for actuator safety. We penalize the negative power beyond a predefined deadband to suppress large regenerative braking events for each joint:
\begin{equation}
\mathcal{L}_{\text{neg-power}}
=
\sum_{j \in \mathcal{J}}
\left(
\frac{
\max(-P_j - P_{\text{db}}, 0)
}{K}
\right)^{2},
\end{equation}
where $P_j$, $P_{\text{db}}$ denotes power for joint $j$ and the deadband threshold, respectively. $K$ is a normalizing constant. In practice, this term is selectively applied to the knee joints in the context of high-dynamic motions(\eg, backflips), as these joints are particularly prone to high braking loads during impacts and recovery phases.

\paragraph{Actuation-Aware Torque--Speed Constraints}
A major source of sim-to-real discrepancy stems from the oversimplification of actuator modeling, whereas standard torque clipping techniques neglect velocity-dependent constraints imposed by back-electromotive force and physical power limits. This omission leads to a significant sim-to-real gap when performing high-dynamic motions. To bridge this gap, we integrate a realistic torque-speed operating envelope directly into the simulation, dynamically deriving torque limits based on the instantaneous alignment of torque and angular velocity:

% lies in actuator modeling. Many prior approaches rely on simple torque clipping, implicitly assuming constant torque availability regardless of joint velocity. This ignores the electromechanical limits imposed by back-electromotive force and power constraints, ultimately leading to a significant sim-to-real gap and catastrophic failures in high-dynamic motions when deployed on physical hardware.
% % , which ignores electromechanical limits imposed by back-electromotive force and power constraints.

% We instead integrate a torque--speed operating envelope directly into the simulation. The base torque limit is first selected based on the relative direction of torque and velocity:
\begin{equation}
\tau_{\max,0} =
\begin{cases}
\tau_{y1}, & v \cdot \tau_{\text{in}} > 0, \\
\tau_{y2}, & v \cdot \tau_{\text{in}} \le 0 .
\end{cases}
\end{equation}

The admissible torque is then defined as a monotonically decreasing function of joint velocity magnitude:
\begin{equation}
\tau_{\text{clipped}}(v) =
\begin{cases}
\tau_{\max,0}, & |v| < v_{x1}, \\
\tau_{\max,0} \left( 1 - \dfrac{|v| - v_{x1}}{v_{x2} - v_{x1}} \right),
& v_{x1} \le |v| \le v_{x2}, \\
0, & |v| > v_{x2}.
\end{cases}
\end{equation}
The commanded torque is finally clipped to this admissible range before being applied to the joint, which ensures that the simulator never samples torque commands that are physically unattainable for the real actuators.

In addition to torque-speed limits, we further model actuator-level internal losses
through a nonlinear friction term applied after torque clipping,
\begin{equation}
\tau_{\text{applied}} =
\tau_{\text{clipped}}
- \left(
\mu_s \tanh\!\left( \frac{v}{v_{\text{act}}} \right)
+ \mu_d v
\right).
\end{equation}
The smoothed Coulomb component captures the transition from static to kinetic friction,
while the viscous term accounts for velocity-dependent dissipation and provides
additional damping. The parameters $\mu_s$, $v_{\text{act}}$, and $\mu_d$ are constants.

Overall, this structured refinement stage yields controllers that are simultaneously safer, more robust to large disturbances, and more faithfully aligned with real-world actuator dynamics, thereby enabling reliable deployment on robots.

% To bridge the Sim-to-Real gap, we implemented more aggressive domain randomization by increasing noise levels, push frequencies, and introducing complex terrains. Furthermore, we integrated actuator-specific corrections—including precise motor modeling—to minimize the discrepancy between simulation and physical deployment.

% \cref{tab:randomization_noise_params} details the rigorous domain randomization and noise configurations utilized during the residual policy refinement phase. Compared to the moderate settings used for base policy pre-training, this "harsh" configuration significantly expands the disturbance envelope to bridge the gap between simulation and unrefined real-world physics. Key enhancements include the enforcement of a 10--15 ms action delay and the introduction of restitution coefficients and extra ankle calibration noise to simulate hardware imperfections. We also intensify initial state noise and external push magnitudes (e.g., angular velocity noise increased by 50\%) to challenge the controller's recovery capabilities. Notably, we introduce terrain surface noise and vertical steps, forcing the agent to adapt to non-flat, irregular ground geometries. Crucially, the Termination Thresholds are scaled to 1.5$\times$ the base values (e.g., orientation error increased to 1.20 rad); this relaxation allows the residual policy to explore and learn recovery behaviors in near-failure states that would otherwise be prematurely terminated, thereby fostering superior robustness.
% \input{tables/table_method_harsh_noise}

\subsection{Real World Deployment}
The integrated real-world deployment pipeline is illustrated in \cref{fig:method_overview}(c). In the deployment phase, we leverage the pelvis IMU as the primary orientation source and compute the torso rotation through \ac{fk}. To ensure minimal control latency, the entire computational pipeline—including \ac{fk}-based state estimation, the base flow matching policy, and the residual policy—is optimized and executed via TensorRT. This integrated pipeline achieves an end-to-end inference latency of about 10ms on the Unitree G1’s onboard Orin NX. Such optimization enables the robot to execute high-quality motion tracking at a consistent 50Hz frequency in complex physical environments.
\section{Experiments}
We present extensive experiments in simulation and on physical robots to evaluate the scalability of our proposed \model system as motion libraries grow in diversity and difficulty. Our experiments are organized around the following key questions:

\noindent\textbf{Q1: Scalable high-fidelity tracking.}
Compared to prior multi-motion baselines, can our approach maintain high-fidelity tracking at scale, both in simulation and on real robots, without collapsing under representation and optimization challenges?

\noindent\textbf{Q2: Fidelity--scalability trade-off (\model v.s. from-scratch RL controllers).}
As motion diversity and difficulty increase, how does tracking performance
degrade for from-scratch multi-motion reinforcement learning controllers, and to what extent can our approach extend the scalability frontier?

\noindent\textbf{Q3: Capacity scaling with flow-based  (\model v.s. MLP-based controllers).}
Does increasing model capacity improve large-scale multi-motion tracking performance, and does generative pretraining via flow matching enable stronger and more stable scaling behavior than conventional MLP-based motion tracking controllers?

\noindent\textbf{Q4: Real-world executability and robustness.}
How do aggressive domain randomization, actuation-aware modeling, and power-aware safety mechanisms individually and jointly affect sim-to-real transfer and real-world execution success?

\noindent\textbf{Q5: Qualitative whole-body capability.}
Beyond scalar tracking metrics, can \model demonstrate agile and versatile whole-body behaviors across diverse motion styles and dynamic contact patterns?

Together, these questions probe the scalability and robustness of \model by disentangling the roles of generative pretraining for representation and capacity scaling, and residual post-training for real-world executability.
\subsection{Experimental Setup}
\subsubsection{Motion Libraries}
We construct our motion libraries following a two-tier design.
First, we use the full \ac{lafan1}~\cite{harvey2020robust}, which has been widely adopted in prior
multi-motion tracking work and serves as a standard benchmark for evaluating scalability
under stylistic and temporal diversity.

Second, to evaluate and push the limit of extreme humanoid motions, we curate about 60 highly challenging motions selected from \ac{lafan1}~\cite{harvey2020robust}, AMASS~\cite{mahmood2019amass}, MimicKit~\cite{peng2025mimickit}, and Reallusion~\cite{reallusion}.
These motions exhibit substantially higher dynamic intensity, frequent contact
transitions, and tight timing constraints, as shown in~\cref{fig:figure1}(a). We collectively refer to this curated set as the \dataset dataset. 

Together, \ac{lafan1} and \dataset form a motion library that spans both standard
multi-motion benchmarks and extreme behaviors that probe the limits of fidelity,
robustness, and real-world executability.

\subsubsection{Baselines}
We compare against two families of strong baselines designed for multi-motion tracking.

\noindent\textbf{(a) Specialist-to-Unified MLP Distillation.}
This class of methods~\cite{zhang2025track} first trains per-motion
(or per-cluster) specialist policies and then distills them into a single unified MLP
tracking policy. Relying on supervised distillation, they benefit from relatively stable and
straightforward optimization, but are limited by the representational
capacity of the MLP policy.

\noindent\textbf{(b) From-scratch Multi-motion Reinforcement Learning.}
This class~\cite{he2024omnih2o,chen2025gmt,ji2024exbody2,zhang2025track,luo2025sonic,liao2025beyondmimic} directly trains a single unified tracking policy from scratch using reinforcement learning across all motions, but often suffers from gradient interference and conservative averaging as motion diversity and difficulty increase.

% \paragraph{\model\textbf{ (Generative Pretraining + Residual RL)} 

% \paragraph{\model(Generative Pretraining + Residual RL)} 
% % \noindent\model\textbf{ (Generative Pretraining + Residual RL).}
% We first distill motion specialists into a unified generative tracking policy using
% flow matching, yielding a scalable tracking prior with high representational capacity.
% We then apply a residual reinforcement learning post-training stage to align the policy
% with real-world actuation constraints and harsh randomization, improving physical executability and robustness
% under deployment.

\subsection{Evaluation Metrics}

% We evaluate motion tracking using pose-based and physics-based metrics in simulation.
The policy is evaluated through simulated rollouts of motion tracking to extract performance metrics. The primary metric is the \textbf{success rate (Succ)}, where an episode is deemed
unsuccessful if the humanoid deviates beyond a predefined threshold from the reference
motion or becomes unstable.
We additionally report the \textbf{root-relative mean per-joint position error (MPJPE)}
(mm), as well as discrepancies in joint-space \textbf{velocity} ($\Delta$vel) and
\textbf{acceleration} ($\Delta$acc), to quantify kinematic accuracy and physical fidelity.

On physical robots, we evaluate performance using deployment-oriented metrics, including \textbf{skill-level
success rates} and
qualitative assessments of motion fidelity for high-dynamic behaviors.

% \subsection{Results and Analyses}
\subsection{Scalable high-fidelity tracking (Q1)}
\label{sec:q1_main}
\begin{table*}[!t]
\vspace{-2mm}
\centering
\small
\setlength{\tabcolsep}{4.5pt}
\renewcommand{\arraystretch}{1.15}
\label{table:q1}
\caption{\small \textbf{Scalable high-fidelity motion tracking under diverse motion sets.}
\model consistently achieves lower kinematic errors and higher success rates than
baselines, particularly on high-dynamic and unseen motions.}
\label{tab:q1_main}

\resizebox{\textwidth}{!}{
\begin{tabular}{l|cccc|cccc|cc}
\hline
Method
& \multicolumn{4}{c|}{LaFAN1+XtremeMotion}
& \multicolumn{4}{c|}{XtremeMotion}
& \multicolumn{2}{c}{Unseen Motions} \\
\cline{2-11}
& MPJPE $\downarrow$ & $\Delta$vel $\downarrow$ & $\Delta$acc $\downarrow$ & Succ.(\%) $\uparrow$
& MPJPE $\downarrow$ & $\Delta$vel $\downarrow$ & $\Delta$acc $\downarrow$ & Succ.(\%) $\uparrow$
& MPJPE $\downarrow$ & Succ.(\%) $\uparrow$ \\
\hline
From-scratch RL~\cite{liao2025beyondmimic}
& 47.95 & 10.03 & 3.27 & 82.95
& 54.19 & 14.04 & 4.04 & 79.45
& 56.87 & 85.29 \\
Specialist$\rightarrow$Unified MLP~\cite{zhang2025track}
& 33.35 & 6.70 & 2.11 & 94.91
& 43.43 & 11.38 & 2.51 & 89.22
& 58.94 & 85.95 \\
\hline
\textbf{OmniXtreme (Pretrain only)}
& 32.65 & 6.34 & \textbf{2.04} & 97.17
& 37.11 & 10.46 & \textbf{2.39} & 95.16
& 56.25 & 89.23 \\
\textbf{OmniXtreme (Pretrain + Post-train)}
& \textbf{30.93} & \textbf{6.19} & 2.13 & \textbf{98.54}
& \textbf{36.17} & \textbf{9.94} & 2.58 & \textbf{95.64}
& \textbf{56.05} & \textbf{89.54} \\
\hline
\end{tabular}}
\vspace{-2mm}
\end{table*}

\begin{table}[t]
\vspace{-1mm}
\centering
% \footnotesize
\small
\setlength{\tabcolsep}{4pt}
\renewcommand{\arraystretch}{1.05}
\caption{\small \textbf{Real-world evaluation of~\model on Unitree G1.}
We evaluate \model on physical hardware using motions drawn from
the \dataset motion library.}
\label{tab:hw_overall}

\begin{tabular}{l|c c c}
\hline
Skill 
& \#Motions 
& Attempts 
& Success (\%)$\uparrow$ \\
\hline
Flip 
& 7 & 55 & 96.36 \\

Handspring 
& 5 & 35 & 88.57 \\

Acrobatics 
& 4 & 15 & 80.00 \\

Breakdance 
& 5 & 22 & 86.36 \\

Martial arts 
& 3 & 30 & 93.33 \\

\hline
\textbf{Total}
& \textbf{24}
& \textbf{157}
& \textbf{91.08} \\
\hline
\end{tabular}

\vspace{-1mm}
\end{table}

In this section, we study whether high-fidelity humanoid motion tracking can be
preserved by \model as motion libraries scale in diversity and difficulty.
We compare \model with specialist-to-unified MLP distillation and from-scratch
multi-motion reinforcement learning under matched model capacity and identical
training data.
All methods are trained on the same combined motion library
(\ac{lafan1} + \dataset) and evaluated on three test sets:
the full motion library, the high-dynamic \dataset subset, and an unseen motion set
(randomly sampled from retargeted AMASS).

\textbf{Simulation.}
As summarized in \cref{tab:q1_main}, \textbf{\model consistently outperforms both baselines
across all simulation metrics.} The gap becomes substantially larger
on \dataset and unseen motions, where baseline methods exhibit reduced success rates
and increased tracking errors as motion difficulty increases. This indicates that \model preserves tracking fidelity as motion diversity and
difficulty scale, rather than degrading under increased complexity.

% \paragraph{Real-world evaluation}
% \noindent\textbf{Real-world evaluation}
\textbf{Real world.}
We further deploy \model on a Unitree G1 humanoid robot using
motions drawn from \dataset. For clarity of presentation, motions are grouped into representative skill categories based on shared dynamic structure and contact patterns. A trial is considered successful if the motion is executed without manual
intervention or safety-triggered termination.
As shown in \cref{tab:hw_overall}, across 157 real-world trials spanning 24 distinct
high-dynamic motions,  \textbf{\model achieves consistently high success rates across diverse
skill categories}, including flips, acrobatics, breakdancing, and martial-arts-style
motions.
These results demonstrate that the scalability gains observed in simulation translate
to robust and physically executable behaviors on real hardware.

\subsection{Fidelity--scalability trade-off (Q2)}
\label{sec:q2_tradeoff}

To characterize the fidelity--scalability trade-off in multi-motion tracking, we progressively scale motion diversity by training on an expanding set of motions
drawn from \dataset, and analyze how different training paradigms respond under the
same evaluation protocol.

\begin{figure}[t]
\vspace{-2mm}
\centering
\includegraphics[width=.4\textwidth]{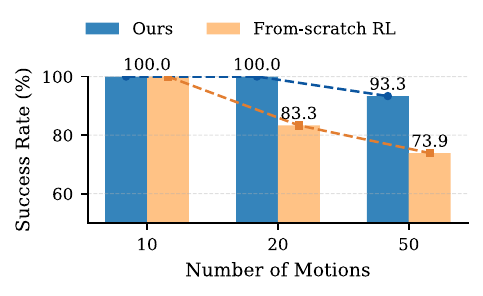}
% \fbox{\rule{0pt}{1.2in}\rule{0.95\linewidth}{0pt}}
\vspace{-2mm}
\caption{\small \textbf{Fidelity--scalability trade-off.}
Tracking success rate as we progressively scale motion diversity and difficulty,
while evaluating all policies on a fixed set of the first 10 motions.}
\label{fig:q2_tradeoff}
\vspace{-2mm}
\end{figure}

Under this controlled scaling regime, \textbf{from-scratch multi-motion reinforcement learning
exhibits earlier and more pronounced performance degradation as scale increases,
whereas \model maintains higher tracking robustness over a broader scaling range. }As shown in \cref{fig:q2_tradeoff}, from-scratch multi-motion RL~\cite{liao2025beyondmimic}
exhibits a characteristic degradation pattern as motion diversity increases:
tracking precision deteriorates steadily, followed by a sharp loss of robustness beyond
a critical scale. These results indicate that the observed fidelity--scalability trade-off is not inherent,
but can be substantially alleviated by a more scalable training paradigm.
% once highly dynamic motions are introduced.
% Failures concentrate on timing-sensitive phases such as takeoff, landing, and rapid contact transitions, and beyond a critical scale, performance collapses across the entire motion set.

\begin{figure}[t]
\vspace{-2mm}
\centering
\includegraphics[width=.47\textwidth]{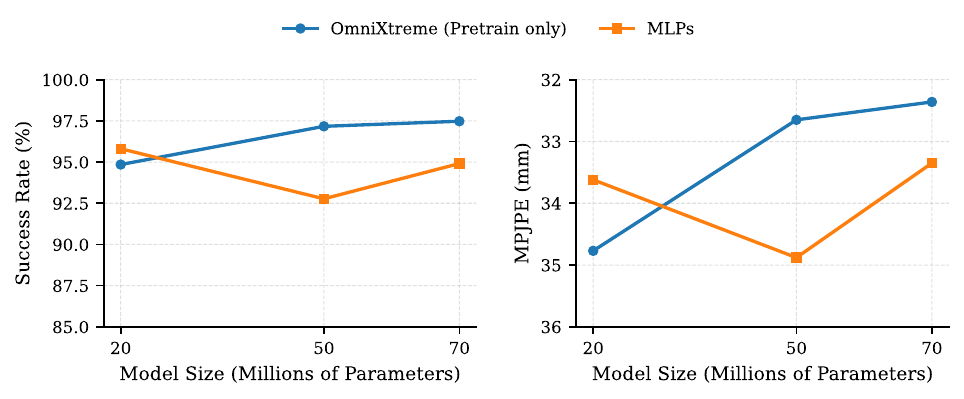}
% \fbox{\rule{0pt}{1.2in}\rule{0.95\linewidth}{0pt}}
\vspace{-2mm}
\caption{\small \textbf{Capacity scaling.}
Tracking fidelity and robustness as a function of model capacity.
\model benefit more strongly from scaling, while conventional MLP controllers saturate earlier.}
\label{fig:q3_capacity}
% \vspace{-2mm}
\vspace{-17pt}
\end{figure}
\subsection{Capacity scaling (Q3)}
\label{sec:q3_capacity}

We next examine whether increasing model capacity further improves multi-motion tracking performance, and whether our generative policy exhibits stronger scaling behavior than conventional MLP-based controllers~\cite{pan2025much}.
We train a family of models with increasing capacity (e.g., width/depth or Transformer hidden size and layers) under the same data and training recipe.
\cref{fig:q3_capacity} reports tracking fidelity and robustness as a function of model capacity.
We observe that \textbf{additional capacity translates more directly into improved tracking quality for our flow matching policies, whereas MLP-based policies show weaker gains.}These results suggest that representational scaling is a practical lever for extending multi-motion tracking fidelity when paired with a scalable training paradigm.

\subsection{Real-world executability and robustness (Q4)}
\label{sec:q4_posttrain}
We analyze the contribution of different post-training mechanisms to sim-to-real
transfer by incrementally enabling them and evaluating real-world execution at the
skill level. \cref{tab:q4_ablation} summarizes the ablation results.

In summary, \textbf{different classes of high-dynamic motions exhibit distinct failure modes,
and each execution-oriented mechanism addresses a complementary aspect of real-world
executability}. For highly impulsive motions such as flips, enforcing actuator torque-speed constraints is sufficient to enable stable execution, as respecting motor envelopes prevents
immediate hardware-level violations. Contact-rich skills such as breakdance and acrobatic motions remain unstable under motor constraints alone, but benefit substantially from aggressive domain randomization, which improves robustness to timing-sensitive contact
perturbations. Motions involving high-speed impact buffering, such as acrobatic landings,
remain challenging even with aggressive domain randomization, power-safety regularization is critical for
these skills, as it mitigates failures caused by excessive transient braking loads
and unsafe energy absorption during high-impact contacts.
Together, these results show that reliable real-world execution emerges from the
combined effects of actuation-aware modeling, robustness-oriented randomization, and
energy-aware safety constraints.

\begin{table}[t]
\vspace{-1mm}
\centering
\small
\setlength{\tabcolsep}{5pt}
\renewcommand{\arraystretch}{1.2}
\caption{\small \textbf{Ablation of post-training mechanisms.}
Real-world executability of different skills under incremental post-training mechanisms.
\textbf{None}: base pretrained policy only;
\textbf{MC}: motor constraints;
\textbf{ADR}: aggressive domain randomization;
\textbf{PS}: power-safety regularization (overcurrent / regenerative protection).
{\color{goodgreen}$\checkmark$}: stable execution;
{\color{midorange}$\triangle$}: unstable or inconsistent execution;
{\color{badred}$\times$}: consistent failure;
{\color{badred}\psicon}: failures primarily associated with power-safety protection,
such as overcurrent or excessive regenerative braking.}
\label{tab:q4_ablation}

\begin{tabular}{l|c c c c}
\hline
Skill 
& None 
& +MC 
& +MC+ADR 
& Full (+MC+ADR+PS) \\
\hline
Flip         
& {\color{midorange}$\triangle$} 
& {\color{goodgreen}$\checkmark$} 
& {\color{goodgreen}$\checkmark$} 
& {\color{goodgreen}$\checkmark$} \\

Breakdance   
& {\color{midorange}$\triangle$} 
& {\color{midorange}$\triangle$} 
& {\color{goodgreen}$\checkmark$} 
& {\color{goodgreen}$\checkmark$} \\

Acrobatics   
& {\color{badred}$\times$} 
& {\color{midorange}$\triangle$} 
& {\psicon} 
& {\color{goodgreen}$\checkmark$} \\
\hline
\end{tabular}
% \vspace{-2mm}
\vspace{-16pt}
\end{table}

\subsection{Qualitative results on extreme motions (Q5)}
\label{sec:q5_qualitative}

\begin{figure}[t]
% \vspace{-2mm}
\centering
\includegraphics[width=.47\textwidth]{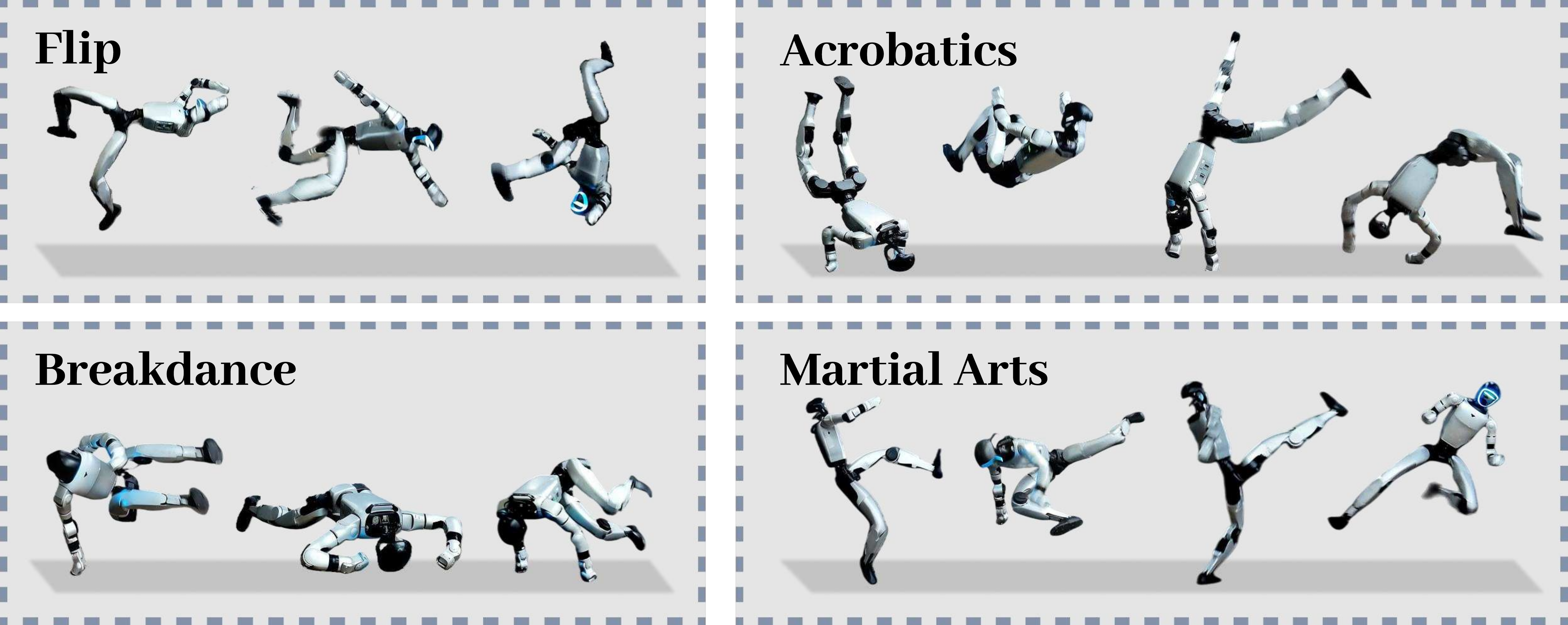}
\vspace{-2mm}
\caption{\small \textbf{Qualitative results.}
Representative real-world rollouts produced by \model, executing qualitatively distinct whole-body motions across diverse styles and contact patterns, including flips, acrobatics, breakdance, and martial-arts behaviors. The results illustrate stable and coordinated execution under rapid contact transitions and timing-sensitive phases on physical hardware.}
\label{fig:q5_qualitative}
% \vspace{-2mm}
\vspace{-17pt}
\end{figure}

Finally, we provide qualitative evidence that \model can exhibit agile and versatile whole-body skills across distinct motion styles and contact patterns, beyond what is captured by scalar tracking metrics.
We visualize representative rollouts spanning stylistic motions from \dataset. \cref{fig:q5_qualitative} highlights that \model can track qualitatively different motions with coherent whole-body coordination, complement the quantitative metrics in Q1-Q4 and illustrate the breadth of behaviors enabled by scalable generative pretraining and actuation-aware refinement. Please refer to the \textit{Supp.} for additional qualitative results, including video demonstrations.

\section{Conclusion} 
\label{sec:conclusion}
We presented \model, a two-stage framework for scalable high-fidelity humanoid motion tracking in high-dynamic regimes. By combining specialist-to-unified flow-based pretraining with actuation-aware residual reinforcement learning, \model mitigates both the learning bottleneck at scale and the physical executability bottleneck at sim-to-real deployment. Extensive simulation results show that \model preserves tracking fidelity substantially deeper into motion diversity than other baselines, and real-robot experiments demonstrate reliable execution of diverse extreme behaviors with a single unified policy, breaking the conventional fidelity--scalability trade-off.

For future research, jointly scaling data diversity and model capacity will be essential for enhancing the generalization of whole-body humanoid motor skills. As learning-based controllers are pushed toward more dynamic and hardware-constrained regimes, actuation-aware modeling becomes a critical component of the learning pipeline. By incorporating high-fidelity actuation characteristics—such as current, power, torque, and speed-dependent constraints—researchers can further bridge the sim-to-real gap, ensuring that learned behaviors translate seamlessly to physical humanoid robots.
\section*{Acknowledgments}

%% Use plainnat to work nicely with natbib. 

\bibliographystyle{plainnat}
\bibliography{reference_header,references}

\clearpage
\section*{Appendix}
\label{sec:appendix}
% detail expression for observation and termination ??? detail of sampling??? but we do not explore well on base sampling??? detail of kp,kd calculation? termination calculation?? reward calculation and detail,but we donot have too much details of observation signals???

%  and experiment batch size, max episode length, ppo configs,rslrl setting???

%  reallusion motion source?

\subsection{\dataset Motion Library}
The challenging motions selected from the AMASS dataset, Reallsusion, MimicKit, and \ac{lafan1} are detailed in \cref{tab:motion_info}.
\begin{table*}[htbp]
  \centering
  \caption{Motion information summary.} % 表格标题
  \label{tab:motion_info}
  \begin{tabularx}{\textwidth}{llX}
    \toprule
    Motion ID & Motion Source & Motion Description \\
    \midrule
    1 & CMU-85-05 & Handstand walk. \\
    2 & CMU-85-10 & Handstand spin. \\
    3 & CMU-88-09 & Back Handspring with a full twist. \\
    4 & CMU-88-08 & Back Handspring with a half twist. \\
    5 & CMU-90-08 & Aerial Cartwheel \\
    6 & CMU-85-14 & B-boying with a rapid-fire string of back handsprings. \\
    7 & CMU-85-13 & Skip back, pivot through a hand-to-headstand, drop, flip, and bounce back. \\
    8 & CMU-90-06 & Fly kick. \\
    9 & CMU-90-34 & Forward roll. \\
    10 & CMU-90-01 & Backward roll. \\
    11 & CMU-90-28 & Backspin. \\
    12 & CMU-85-08 & Thomas Flare. \\
    13 & CMU-85-12 & Long breaking dance. \\
    14 & CMU-85-04 & Another long breaking dance. \\
    15 & CMU-85-01 & Bicycle kick flip. \\
    16 & CMU-85-02 & Another bicycle kick flip. \\
    17 & CMU-88-06 & Butterfly kick. \\
    18 & CMU-85-06 & Webster flip. \\
    19 & CMU-49-08 & Two consecutive cartwheels. \\
    20 & CMU-90-30 & Alternating pistol squats. \\
    21 & CMU-90-29 & Acrobatic gymnastics with cartwheel and back handsprings. \\
    22 & CMU-90-19 & Crawl forward and backflip. \\
    23 & GeneralA11-MilitaryCrawlForward & Stay low and crawl forward. \\
    24 & IconicHeroMotion-SwordJudgment & Spinning slash. \\
    25 & IconicHeroMotion-SwordHeroic & Another style of spinning slash. \\
    26 & HandtoHandCombat-B3AttackReverseTurningKick & Reverse turning kick. \\
    27 & HandtoHandCombat-D2AttackPunchSweepKick & Punch and sweep kick. \\
    28 & HandtoHandCombat-D4AttackReverseFrontSnapKick & Reverse and front snap kick. \\
    29 & HandtoHandCombat-D4DodgeRollBack & Two consecutive rolls in different styles. \\
    30 & HandtoHandCombat-G1GetupKipUp & Kip up. \\
    31 & HandtoHandCombat-G2GetupHandstandKipUp & Handstand kip. \\
    32 & HandtoHandCombat-KO2FalltoGroundAxelDown & Execute a downward Axel into a ground fall. \\
    33 & LadyAgent-AgentElbowStrikeSweepKick & Elbow strike and sweep kick. \\
    34 & LadyAgent-AgentHandspring & Front handspring. \\
    35 & LadyAgent-AgentRollForward & Shoulder roll. \\
    36 & LadyAgent-AgentShootSidewardRoll & Quick side roll. \\
    37 & LadyAgent-AgentSnapKick & Snapkick. \\
    38 & Mimickit-g1spinkick & Spinkick. \\
    39 & \ac{lafan1}-dance1subject2 [82.8,106.9]s & Constantly spin in full circles. \\
    40 & \ac{lafan1}-dance1subject2 [145.3,161.3]s & Play guitar while hopping on one leg. \\
    41 & \ac{lafan1}-dance2subject3 [160.2,224.3]s & Flutter arms and  hands. \\
    42 & \ac{lafan1}-fightandsports1subject1 [167.0,176.9]s & Continuous long jumps. \\
    43 & \ac{lafan1}-fightandsports1subject1 [0.0,17.0]s & Balance on one leg. \\
    44 & \ac{lafan1}-dance1subject1 [104.6,119.1]s & Cartwheel twice. \\
    45 & \ac{lafan1}-jump1subject1 [70.3,87.6]s & Play hopscotch. \\
    46 & \ac{lafan1}-jump1subject1 [89.1,138.4]s & Hop on one foot. \\
    47 & \ac{lafan1}-jump1subject1 [0.0,72.0]s & Successive leaps. \\
    48 & \ac{lafan1}-fightandsports1subject4 [154.4,219.8]s & Vigorously swing the baseball bat. \\
    49 & \ac{lafan1}-fightandsports1subject4 [61.1,85.9]s & Lash the golf club. \\
    50 & \ac{lafan1}-fightandsports1subject4 [28.2,61.9]s & Diverse kicking movements. \\
    51 & \ac{lafan1}-run1subject2 [20.4,101.0]s & Shuttle run. \\
    52 & \ac{lafan1}-run1subject2 [92.0,130.6]s & Run rapidly. \\
    53 & \ac{lafan1}-fallandgetup2subject3 [33.8,56.1]s & Kip up twice. \\
    54 & \ac{lafan1}-fightandsports1subject1 [17.3,27.2]s & Roundhouse kick. \\
    55 & \ac{lafan1}-jump1subject2 [187.9,196.2]s & Push up. \\
    56 & \ac{lafan1}-jump1subject2 [196.0,205.8]s & Lateral roll. \\
    57 & \ac{lafan1}-jump1subject2 [205.6,244.4]s & Lateral roll and kip up. \\
    \bottomrule
  \end{tabularx}
\end{table*}
% Motion-complexity metrics used for comparing motion datasets.
% This file is intended to be \input{} into the main paper.

\subsection{Motion Complexity Metrics}
\label{sec:motion-metrics}

Given a motion sequence with frame index $t \in \{0,\dots,T-1\}$, frame time-step $\Delta t$, joint positions $q_t \in \mathbb{R}^J$, joint velocities $\dot{q}_t \in \mathbb{R}^J$, root/body positions $x_t \in \mathbb{R}^{B\times 3}$ and velocities $v_t \in \mathbb{R}^{B\times 3}$, we compute a set of scalar metrics that summarize its kinematic and contact complexity.

\paragraph{Joint/base kinematics.}
From joint or base kinematics we extract:
\begin{align}
v_{\max} &= \max_{t,j} \;\bigl\lVert \dot{q}_t^{(j)} \bigr\rVert , \\
a_{\max} &= \max_{t,j} \;\bigl\lVert \ddot{q}_t^{(j)} \bigr\rVert , \\
j_{\max} &= \max_{t,j} \;\bigl\lVert \dddot{q}_t^{(j)} \bigr\rVert ,
\end{align}
where $\ddot{q}_t$ and $\dddot{q}_t$ are obtained by finite differencing,
e.g.,
\(
 \ddot{q}_t \approx (\dot{q}_{t+1} - \dot{q}_{t}) / \Delta t
\)
and
\(
 \dddot{q}_t \approx (\ddot{q}_{t+1} - \ddot{q}_{t}) / \Delta t
\).
We use base (torso) linear and angular kinematics to define the same three scalars (maximum linear velocity, linear acceleration and angular velocity).

\paragraph{Center-of-mass vertical velocity.}
Let $z_t^{\mathrm{CoM}}$ be the center-of-mass height and $\dot{z}_t^{\mathrm{CoM}}$ its vertical velocity.
We measure the peak vertical CoM speed
\begin{equation}
 v^{\mathrm{CoM}}_{z,\max} = \max_t \bigl\lvert \dot{z}_t^{\mathrm{CoM}} \bigr\rvert .
\end{equation}
This captures the dynamic “liveliness'' of the motion (e.g., jumps).

\paragraph{Airborne ratio.}
Given a set of foot body indices $\mathcal{F}$, a frame is considered \emph{airborne} if all feet
are above a small height threshold $h_{\text{air}}$ above the ground. The airborne ratio is
the fraction of airborne frames
\begin{equation}
 \mathrm{Airborne} = \frac{1}{T} \sum_{t=0}^{T-1}
   \mathbf{1}\!\left[
     \min_{b \in \mathcal{F}} z_t^{(b)} > h_{\text{air}}
   \right],
\end{equation}
where $z_t^{(b)}$ denotes the vertical position of body $b$ at time $t$.

\paragraph{Contact switch frequency.}
Independently, we obtain a binary contact state $c_t \in \{0,1\}^K$ for $K$ end-effectors
and define the contact switch frequency as the number of contact state flips per second:
\begin{equation}
 f_{\text{switch}} =
  \frac{1}{(T-1)\,\Delta t}
  \sum_{t=0}^{T-2} \mathbf{1}[c_{t+1} \neq c_t] \;\;[\text{Hz}].
\end{equation}
In the radar plots, the \emph{contact switch} axis corresponds to a normalized version of this
scalar, $s_{\text{sw}} = \min(f_{\text{switch}}/10, 1)$.

\paragraph{Difficulty scores.}
For comparability across metrics with different physical units we map the main scalar metrics to $[0,1]$ “difficulty'' scores using simple clamping and linear scaling,
mirroring the implementation used in our analysis code.
Let
\[
\begin{aligned}
s_{ang}   &= \min\!\left(\frac{ang_{\max}}{20},\,1\right), \\
s_v       &= \min\!\left(\frac{v_{\max}}{20},\,1\right), \\
s_a       &= \min\!\left(\frac{a_{\max}}{200},\,1\right), \\
s_{\text{air}} &= \mathrm{clip}\!\bigl(\mathrm{Airborne},\,0,\,1\bigr), \\
s_{\text{com}} &= \min\!\left(\frac{v^{\mathrm{CoM}}_{z,\max}}{2},\,1\right), \\
s_{\text{sw}}  &= \min\!\left(\frac{f_{\text{switch}}}{10},\,1\right),
\end{aligned}
\]

where $\mathrm{clip}(x,0,1)=\max(0,\min(x,1))$.
The resulting 6-D score vector
\begin{equation}
 \mathbf{s} = \bigl[s_{ang},\,s_v,\,s_a,\,s_{\text{com}},\,s_{\text{air}},\,s_{\text{sw}}\bigr]
\end{equation}
is what we use in the radar plots to summarize and compare motion complexity across datasets.

\subsection{Teacher Training}
The teacher policy is trained using BeyondMimic~\cite{liao2025beyondmimic}——a RL motion tracking algorithm within the IsaacLab simulator, with the specific reward constituents detailed in \cref{tab:reward_params_teacher}. Regarding the robot configuration, a full-mesh representation is employed for the collision geometry to ensure high-fidelity physical interactions. System parameters—including armature, $k_p$, $k_d$, and action scale—are kept consistent with the BeyondMimic framework, which is specifically tailored to the motor specifications of each joint.

The Teacher Policy employs an MLP architecture with hidden dimensions of [512, 256, 128] for both actor and critic. The actor observation is defined by
\[
\mathbf{o}=[\boldsymbol{\psi},\,\mathbf{e}_{\text{torso}},\,\mathcal{V}_{\text{imu}},\,\boldsymbol{q}- \boldsymbol{q}^{0},\,\dot{\boldsymbol{q}},\,\mathbf{a}_{\text{last}}],
\]
including reference motion joint positions and velocities $\boldsymbol{\psi}=[\boldsymbol{q}^{\text{ref}},\,\dot{\boldsymbol{q}}^{\text{ref}}]$, $\mathbf{e}_{\text{torso}}$ which includes torso position difference and torso orientation difference from reference represented by the 6D orientation from $R_{\text{torso}}^{\text{ref}}R_{\text{torso}}^{\top}$, base linear and angular velocities $\mathcal{V}_{\text{imu}}\in\mathbb{R}^{6}$, relative joint positions $\boldsymbol{q}- \boldsymbol{q}^{0}$ computed by joint position $\boldsymbol{q}$ subtract default joint position $\boldsymbol{q}^{0}$ and joint velocities $\dot{\boldsymbol{q}}$, and the previous action $\mathbf{a}_{\text{last}}$, all of which are subject to noise described in \cref{tab:randomization_params_standard_size} to enhance robustness. In contrast, the critic observation includes these same elements in their ground-truth form, supplemented by privileged information such as the robot's precise body position and orientation, providing a comprehensive and noise-free state representation to facilitate stable value function estimation during training.

The detailed training parameters are shown in \cref{tab:ppo_hyperparameters}. To speed up training, early termination is introduced when tracking errors is larger than a threshold. Tracking errors are defined as $\mathbf{e}_{p,b} = \mathbf{p}_{b}^\mathrm{ref} - \mathbf{p}_b$ for position and $\mathbf{e}_{R,b} = \log(R_{b}^\mathrm{ref} R_b^\top)$ for orientation. Termination is triggered if the vertical position error $|\mathbf{e}_{p,z,b}|$ of the torso or any end-effector ($\mathcal{B}_\text{ee}$) exceeds $0.25\text{m}$, or if the local gravity vector discrepancy caused by torso's orientation error $\|\mathbf{e}_{R,\text{torso}}\|$ exceeds $0.8$.
\begin{table}[ht]
  \centering
  \caption{\textbf{Motion tracking hyperparameters for teacher training}~\cite{liao2025beyondmimic}.}
  \label{tab:ppo_hyperparameters}
  \begin{tabular}{lr}
    \\
    \hline
    Hyperparameter & Value \\
    \hline
    \multicolumn{2}{c}{Architecture} \\
    \hline
    Actor MLP hidden dimensions & [512, 256, 128] \\
    Critic MLP hidden dimensions & [512, 256, 128] \\
    Activation function & ELU \\
    \hline
    \multicolumn{2}{c}{Training} \\
    \hline
    Steps per environment & 24 \\
    % Max iterations & 30,000 \\
    % Learning rate & $1 \times 10^{-3}$ \\
    Clip parameter & 0.2 \\
    Entropy coefficient & 0.005 \\
    Value loss coefficient & 1.0 \\
    Discount factor ($\gamma$) & 0.99 \\
    GAE $\lambda$ & 0.95 \\
    Desired KL & 0.01 \\
    Learning epochs & 5 \\
    Mini-batches & 4 \\
    \hline
  \end{tabular}
\end{table}

\begin{table}[t]
\centering
\caption{Reward function terms and expressions for Teacher Policy training~\cite{liao2025beyondmimic}.}
\label{tab:reward_params_teacher}
\begin{tabular}{@{}lcc@{}}
\toprule
\textbf{Reward Term} & \textbf{Weight} & \textbf{Expression} \\ \midrule
Global Torso Position & 0.5 & $\exp(-\|p_{err}\|^2 / 0.3^2)$ \\
Global Torso Orientation & 0.5 & $\exp(-\|\theta_{err}\|^2 / 0.4^2)$ \\
Relative Body Position & 1.0 & $\exp(-\|p_{rel\_err}\|^2 / 0.3^2)$ \\
Relative Body Orientation & 1.0 & $\exp(-\|\theta_{rel\_err}\|^2 / 0.4^2)$ \\
Body Linear Velocity & 1.0 & $\exp(-\|v_{err}\|^2 / 1.0^2)$ \\
Body Angular Velocity & 1.0 & $\exp(-\|\omega_{err}\|^2 / 3.14^2)$ \\
Action Rate & -0.1 & $\|a_{t} - a_{t-1}\|^2$ \\
Joint Limit & -10.0 & $\sum \max(0, q - q_{limit})$ \\
Undesired Contacts & -0.1 & $\sum 1(F_{contact} > 1.0)$ \\ \bottomrule
\end{tabular}
\end{table}

% \subsection{Details of Term Definition(expression from beyondmimic, not sure if add it)}
% observation

% The observation space comprises: (i) Motion Phase, consisting of reference joint states $\boldsymbol{\psi}=[\boldsymbol{\theta}^{\text{ref}},\,\dot{\boldsymbol{\theta}}^{\text{ref}}]$ that serve as temporal progress cues rather than direct tracking targets; (ii) Anchor Pose Error $\mathbf{e}_{\text{anchor}}\in\mathbb{R}^{9}$, which stacks the position error $\mathbf{e}_{\mathbf{p},\text{anchor}}$ with a 6D orientation error (Rot6D \cite{zhou2019continuity}) derived from $R_{\text{anchor}}^{\text{des}}R_{\text{anchor}}^{\top}$ to provide global feedback for balance and drift correction; and (iii) Other Proprioception, specifically the IMU twist $\mathcal{V}_{\text{imu}}\in\mathbb{R}^{6}$ expressed in the local IMU frame.

% termination

% Tracking errors are defined as $\mathbf{e}_{p,b} = \mathbf{p}_{b}^\mathrm{d} - \mathbf{p}_b$ for position and $\mathbf{e}_{R,b} = \log(R_{b}^\mathrm{d} R_b^\top)$ for orientation. Termination is triggered if the vertical position error $|\mathbf{e}_{p,z,b}|$ of the anchor or any end-effector ($\mathcal{B}_\text{ee}$) exceeds $0.25\text{m}$, or if the anchor's orientation error $\|\mathbf{e}_{R,\text{anchor}}\|$ exceeds $0.8\text{rad}$

% reward(\cref{tab:rewardterms})

\subsection{Flow matching policy training}
The inputs robot proprioception $p$ and motion command $c$ are each mapped to a single token, while the 15-step history $h$ is mapped to 15 individual tokens, resulting in a concatenated sequence of 17 tokens. Each mapping is performed by a dedicated MLP preprocessing unit consisting of two 256-unit layers with ReLU activations. This sequence is then processed by a Transformer encoder, followed by global average pooling and a linear projection to produce a 1024-dimensional embedding.

Simultaneously, the action $a_t$ is projected through a linear layer to a 1024-dimensional space, while the time step $t$ is transformed into a 1024-dimensional embedding using sinusoidal encoding. These three components—the state embedding, action embedding, and time embedding—are concatenated and fed into a deep MLP with hidden dimensions of $[2048, 2048, 2048]$ to predict the velocity $v_t$ at time $t$.

% \begin{figure}[htbp]
% \centering
% \includegraphics[width=0.995\columnwidth]{appendix/figure/base_white.png}
% % to change color get normal pdf
% % \includesvg[width=.995\columnwidth]{appendix/figure/base2.svg}
% \caption{Overview of the \model's flow matching base model. }
% \label{fig:basemodel}
% \vspace{-17pt}
% \end{figure}
\subsection{Residual policy training}
The Residual Policy is implemented as an MLP with hidden layer sizes of [1024, 512, 256, 128, 64], with the corresponding reward functions detailed in \cref{tab:reward_params}.

The residual learning framework employs an asymmetric actor-critic architecture, as well. In this setup, the residual actor's observation space is restricted to proprioceptive data, relative motion torso orientation, motion command joint pos and joint velocity, the previous step's total action, and the base action out by the flow matching policy. To enhance sim-to-real robustness, uniform noise is injected into these observations, and the noise is aligned with the flow matching base policy noise. 

Conversely, the residual critic utilizes privileged information available only during simulation. The critic's observation space encompasses the full ground-truth state, including noise-free versions of the actor's inputs, as well as critical state variables hidden from the actor, such as the robot's global body position and orientation, base linear velocity, and the precise position of the motion torso.
% \begin{table}[t]
% \centering
% \caption{Reward function terms and expressions for Residual Policy training.}
% \label{tab:reward_params}
% \begin{tabular}{@{}lcc@{}}
% \toprule
% \textbf{Reward Term} & \textbf{Weight} & \textbf{Expression} \\ \midrule
% Global Anchor Position & 0.5 & $\exp(-\|p_{err}\|^2 / 0.3^2)$ \\
% Global Anchor Orientation & 0.5 & $\exp(-\|\theta_{err}\|^2 / 0.4^2)$ \\
% Relative Body Position & 1.0 & $\exp(-\|p_{rel\_err}\|^2 / 0.3^2)$ \\
% Relative Body Orientation & 1.0 & $\exp(-\|\theta_{rel\_err}\|^2 / 0.4^2)$ \\
% Body Linear Velocity & 1.0 & $\exp(-\|v_{err}\|^2 / 1.0^2)$ \\
% Body Angular Velocity & 1.0 & $\exp(-\|\omega_{err}\|^2 / 3.14^2)$ \\
% Action Rate & -0.1 & $\|a_{t} - a_{t-1}\|^2$ \\
% Joint Limit & -10.0 & $\sum \max(0, q - q_{limit})$ \\
% Undesired Contacts & -0.1 & $\sum 1(F_{contact} > 1.0)$ \\
% Power Safety Regularization& xxx  & $xxxxx$ \\
% \bottomrule
% \end{tabular}
% \end{table}

\begin{table}[t]
\centering
\caption{Reward function terms and expressions for Residual Policy training. Refer to \cref{sec:power} for further details regarding power safety regularization.}
\label{tab:reward_params}
\begin{tabular}{@{}lcc@{}}
\toprule
\textbf{Reward Term} & \textbf{Weight} & \textbf{Expression} \\ \midrule
Global Torso Position & 0.5 & $\exp(-\|p_{err}\|^2 / 0.3^2)$ \\
Global Torso Orientation & 0.5 & $\exp(-\|(\theta_{err}\|^2 / 0.4^2)$ \\
Relative Body Position & 1.0 & $\exp(-\|p_{rel\_err}\|^2 / 0.3^2)$ \\
Relative Body Orientation & 1.0 & $\exp(-\|\theta_{rel\_err}\|^2 / 0.4^2)$ \\
Body Linear Velocity & 1.0 & $\exp(-\|v_{err}\|^2 / 1.0^2)$ \\
Body Angular Velocity & 1.0 & $\exp(-\|\omega_{err}\|^2 / 3.14^2)$ \\
Action Rate & -0.1 & $\|a_{t} - a_{t-1}\|^2$ \\
Joint Limit & -10.0 & $\sum \max(0, q - q_{limit})$ \\
Undesired Contacts & -0.1 & $\sum 1(F_{contact} > 1.0)$ \\
Power Safety Regularization& -10.0 &  $\sum \left(\frac{\max\!\bigl(0,\; - \tau \dot{q}- 150\bigr)}{500}\right)^{\!2}$ \\
\bottomrule
\end{tabular}
\end{table}

\subsection{Actuator Modeling}
The physical parameters of the actuators, including armature inertia $I$, torque-speed characteristics, and friction coefficients, are summarized in \cref{tab:actuator_model}.

\begin{table}[htbp]
  \centering
  \setlength{\tabcolsep}{3pt}
  % \footnotesize
  \caption{\textbf{Actuator modeling parameters.}}
  \label{tab:actuator_model}
  \begin{tabular}{lcccccccc}
    \hline
    Actuator & $\tau_{y1}$ & $\tau_{y2}$ & $v_{x1}$ & $v_{x2}$ & $\mu_{s}$ & $v_{act}$ & $\mu_{d}$ & $I$ \\
    \hline
    5020-16   & 24.8  & 31.9  & 30.86 & 40.13 & 0.6 & 0.01 & 0.06 & 3.610e-03 \\
    7520-14.3 & 71.0  & 83.3  & 22.63 & 35.52 & 1.6 & 0.01 & 0.16 & 1.018e-02 \\
    7520-22.5 & 111.0 & 131.0 & 14.5  & 22.7  & 2.4 & 0.01 & 0.24 & 2.510e-02 \\
    4010-25   & 4.8   & 8.6   & 15.3  & 24.76 & 0.6 & 0.01 & 0.06 & 4.250e-03 \\
    \hline
  \end{tabular}
\end{table}

\subsection{PD control}
Following the methodology of BeyondMimic, we calculate the PD gains, $k_p$ and $k_d$, based on the natural frequency ($\omega$), damping ratio ($\zeta$), and the armature inertia ($I$) of the motors:$$k_{\mathrm{p},j} = I_j \omega^2, \quad k_{\mathrm{d},j} = 2 I_j \zeta \omega$$For the configuration addressed here, the parameters consist of a 10 Hz natural frequency and a damping ratio of 2. To determine the action scale, we still utilize the maximum torque values ($\boldsymbol{\tau}_{\text{max}}$) originally defined in the URDF, scaled by a factor of 0.25\cite{liao2025beyondmimic}:
$$\boldsymbol{\alpha} = 0.25\frac{\boldsymbol{\tau}_{\text{max}}}{k_{\mathrm{p},j}}$$ Notably, due to differences in the G1 robot variants, we utilize 7520-22.5 motors for the hip pitch actuators(\cref{tab:joint_motor_mapping}), rather than the 7520-14.3 models used in the original BeyondMimic.

\begin{table}[t]
    \centering
    \caption{Unitree G1 Joint-to-Motor Mapping.}
    \label{tab:joint_motor_mapping}
    \begin{tabular}{ll}
        \toprule
        \textbf{Joint Name} & \textbf{Motor Model} \\ 
        \midrule
        Hip Pitch Joint & 7520-22.5 \\
        Hip Roll Joint & 7520-22.5 \\
        Knee Joint & 7520-22.5 \\
        Hip Yaw Joint & 7520-14.3 \\
        \midrule
        Ankle Pitch Joint & 5020 \\
        Ankle Roll Joint & 5020 \\
        \midrule
        Waist Roll Joint & 5020 \\
        Waist Pitch Joint & 5020 \\
        Waist Yaw Joint & 7520-14.3 \\
        \midrule
        Shoulder Pitch Joint & 5020 \\
        Shoulder Roll Joint & 5020 \\
        Shoulder Yaw Joint & 5020 \\
        Elbow Joint & 5020 \\
        Wrist Roll Joint & 5020 \\
        Wrist Pitch Joint & 4010 \\
        Wrist Yaw Joint & 4010 \\
        \bottomrule
    \end{tabular}
\end{table}

The PD setpoints (target positions) are computed using the following relationship:$$\boldsymbol{q}^{\text{tar}} = \boldsymbol{q}^{0} + \boldsymbol{\alpha} \odot \mathbf{a}$$where $\boldsymbol{q}^{0}$ represents the default joint positions, and $\mathbf{a}$ is the action output by the policy. The pre-clipped control torque is then calculated using the standard PD control law:$$\boldsymbol{\tau} = k_p (\boldsymbol{q}^{\text{tar}} - \boldsymbol{q}) - k_d \dot{\boldsymbol{q}}$$In this expression, $\mathbf{q}$ and $\dot{\mathbf{q}}$ denote the current joint positions and velocities, respectively.

% \subsection{Some Fomular to add?}
% % \subsubsection*{Heuristically Designed Paramters}
%  $k_{\mathrm{p},j} = I_j \omega^2, \quad k_{\mathrm{d},j} = 2 I_j \zeta \omega$, 
% $\boldsymbol{\alpha} = 0.25\frac{\boldsymbol{\tau}_{\text{max}}}{k_{\mathrm{p},j}}$

% $\boldsymbol{\tau}_{j,\text{max}}$

% \[
% \mathbf{o}=[\boldsymbol{\psi},\,\mathbf{e}_{\text{anchor}},\,\mathcal{V}_{\text{imu}},\,\boldsymbol{\theta}- \boldsymbol{\theta}^{0},\,\dot{\boldsymbol{\theta}},\,\mathbf{a}_{\text{last}}],
% \]

% \(\boldsymbol{\psi}=[\boldsymbol{\theta}^{\text{ref}},\,\dot{\boldsymbol{\theta}}^{\text{ref}}]\),

% \(\mathbf{e}_{\text{anchor}}\in\mathbb{R}^{9}\)

% \({\mathbf{e}}_{\mathbf{p},\text{anchor}}\)

% \(R_{\text{anchor}}^{\text{des}}R_{\text{anchor}}^{\top}\)
%  \(\mathcal{V}_{\text{imu}}\in\mathbb{R}^{6}\)

%  $\boldsymbol{\theta}^{\text{sp}} = \boldsymbol{\theta}^{0} + \boldsymbol{\alpha}\odot \mathbf{a}$

\subsection{Evaluation Protocol and Metrics}
\label{sec:supp_eval}

This subsection details the evaluation protocol and metrics used throughout the paper.
Our goal is to assess tracking fidelity, robustness, and generalization under controlled
conditions that are consistent with training and reflective of sim-to-real performance.

\paragraph{Motion segmentation and episode length.}
To avoid bias introduced by varying motion durations, all reference motions are segmented
into fixed-length clips of \textbf{10 seconds} (corresponding to \textbf{500 control steps}),
which matches the episode length used during training.
Motions shorter than 10 seconds are evaluated without further segmentation.
All reported metrics in simulation and real world are computed at the clip level.

\paragraph{Simulation evaluation settings.}
During simulation evaluation, we retain the same sensor noise and base domain randomization
used during training, including center-of-mass offsets and default joint position offsets.
This ensures that evaluation reflects basic sim-to-real robustness rather than idealized
noise-free tracking.
Each motion clip is evaluated over \textbf{10 independent rollouts}, and metrics are averaged
across these trials.

\paragraph{Unseen motion set.}
In addition to test motions drawn from the training distributions, we evaluate generalization
on an unseen motion set.
This set is constructed by uniformly sampling \textbf{1000} motion clips (10 seconds each)
from retargeted AMASS CMU and KIT mocap sequences, explicitly excluding all motions used in
training or selected as extreme motions.
The unseen set primarily consists of locomotion, turning, and simple dance-like behaviors,
and is used to assess generalization beyond the curated extreme motion library.

\paragraph{Evaluation metrics.}
We employ a set of pose-based and physics-aware metrics commonly used in humanoid motion
tracking.
All metrics are computed per episode and then averaged across episodes.

Let $t = 1, \dots, T$ index control steps in an episode with step time $\Delta t$ (seconds),
and let $i = 1, \dots, N$ index tracked bodies.
All per-body errors are computed over the same tracked-body subset.

\paragraph{Root-relative body positions.}
Reference motions are aligned to the robot using the torso pose.
We denote the aligned reference body position as $\mathbf{p}^{\text{ref}}_{t,i}$
and the robot body position as $\mathbf{p}^{\text{rob}}_{t,i}$.

\paragraph{MPJPE (mm).}
\begin{equation}
\mathrm{MPJPE} =
1000 \cdot \frac{1}{T} \sum_{t=1}^{T}
\left(
\frac{1}{N} \sum_{i=1}^{N}
\left\|
\mathbf{p}^{\text{ref}}_{t,i} -
\mathbf{p}^{\text{rob}}_{t,i}
\right\|_2
\right).
\end{equation}

\paragraph{$\Delta$vel (mm/frame).}
Let $\mathbf{v}^{\text{ref}}_{t,i}$ and $\mathbf{v}^{\text{rob}}_{t,i}$ denote linear velocities
(in m/s).
\begin{equation}
\Delta v =
1000 \cdot \Delta t \cdot \frac{1}{T} \sum_{t=1}^{T}
\left(
\frac{1}{N} \sum_{i=1}^{N}
\left\|
\mathbf{v}^{\text{ref}}_{t,i} -
\mathbf{v}^{\text{rob}}_{t,i}
\right\|_2
\right).
\end{equation}

\paragraph{$\Delta$acc (mm/frame$^2$).}
Acceleration is computed by finite differences on velocity:
\[
\mathbf{a}_{t,i} =
\frac{\mathbf{v}_{t,i} - \mathbf{v}_{t-1,i}}{\Delta t}.
\]
The per-step acceleration error is:
\begin{equation}
e^{\text{acc}}_t =
\frac{1}{N} \sum_{i=1}^{N}
\left\|
\mathbf{a}^{\text{ref}}_{t,i} -
\mathbf{a}^{\text{rob}}_{t,i}
\right\|_2.
\end{equation}
We exclude the step immediately following an environment reset.
The reported acceleration error is:
\begin{equation}
\Delta a =
1000 \cdot \Delta t^2 \cdot \frac{1}{T} \sum_{t=1}^{T} e^{\text{acc}}_t.
\end{equation}

\paragraph{Success rate.}
An episode is considered successful if it terminates due to time-out
(i.e., runs for its full allotted duration) rather than early termination.
\textbf{The termination criteria used during evaluation are identical to those used during training},
including violation of safety thresholds.
Success rate is defined as the fraction of successful episodes.

\paragraph{Aggregation.}
MPJPE, $\Delta v$, and $\Delta a$ are first averaged over control steps within each episode.
Final reported metrics are obtained by averaging over all evaluated episodes
(and over motion clips when evaluating multiple motions).

\subsection{Skill-level Grouping for Real-world Evaluation}
\label{sec:supp_skills}

For real-world evaluation, we organize individual motions into a set of
\emph{extreme skill categories}.
Each skill groups multiple motion instances that share similar
semantic intent and high-level dynamic structure.
This grouping is used \emph{only for reporting and analysis}, and does not
affect training or policy execution.

The classification is based on human semantic understanding of the motions,
taking into account characteristic body coordination patterns,
contact configurations, and dominant dynamic features
(e.g., aerial phases, rapid contact switching, or impulsive landing).
We emphasize that this grouping is not obtained through automated clustering
or learned representations, but serves as a transparent and interpretable
abstraction for summarizing real-world performance.
\begin{table}[t]
\vspace{-1mm}
\centering
\footnotesize
\setlength{\tabcolsep}{6pt}
\renewcommand{\arraystretch}{1.15}
\caption{\textbf{Skill-level grouping of real-world evaluation motions.}
Each skill category groups multiple motion instances with similar
semantic meaning and dynamic structure.
Motion IDs correspond to retargeted motions used during hardware evaluation.}
\label{tab:skill_mapping}

\begin{tabular}{l|p{0.68\linewidth}}
\hline
Skill category & Motion IDs \\
\hline
Flip &
5, 9, 10, 15, 16, 18, 22
\\

Handspring &
2, 3, 4, 19, 23
\\

Acrobatics &
1, 6, 7, 20, 21
\\

Breakdance &
11, 12, 13, 14
\\

Martial arts &
8, 17, 38

\\
\hline
\end{tabular}
\vspace{-2mm}
\end{table}

\cref{tab:skill_mapping} lists the correspondence between skill categories
and the underlying motion instances used in hardware evaluation.

\subsection{Motion Subsets Used in Fidelity--Scalability Analysis}
\label{sec:supp_q2_motions}
In the fidelity--scalability analysis~\cref{sec:q2_tradeoff}, we evaluate how tracking performance
degrades as the size of the motion library increases.
To ensure a controlled and interpretable comparison, we explicitly specify
the motion subsets used at each scale.

We begin with a curated set of extreme motions that exhibit
high angular velocities, rapid contact switching, and challenging balance conditions.
These motions represent a diverse yet highly demanding subset of the full motion library.
All quantitative evaluations in Q2 are conducted on the same \textbf{first 10 motions}
of this set, which include fast flips, contact-rich acrobatics, and extreme balance behaviors.
By fixing the evaluation set, we isolate the effect of training-scale growth
from changes in evaluation difficulty.

To study scalability, we progressively expand the training motion library
to include 10, 20, and all extreme motions.
The additional motions are drawn from the same extreme motion pool and increase
diversity in contact patterns and dynamic regimes, while preserving a consistent
difficulty level.
This design yields a controlled yet representative benchmark for analyzing
the fidelity--scalability trade-off under increasingly diverse extreme motions.
\begin{table}[t]
\vspace{-1mm}
\centering
\footnotesize
\setlength{\tabcolsep}{6pt}
\renewcommand{\arraystretch}{1.15}
\caption{\textbf{Motion subsets used in Q2 fidelity--scalability analysis.}
All evaluations are performed on the same first 10 extreme motions.
Larger training sets (20 and all motions) extend this core set with additional
diverse extreme motions.}
\label{tab:q2_motion_ids}

\begin{tabularx}{\linewidth}{c|X}
\hline
Training motions & Motion IDs \\
\hline
10  & 3--10, 13, 14 \\
20  & 2--10, 13-22 \\
50  & All \\
\hline
\end{tabularx}
\vspace{-2mm}
\end{table}

\cref{tab:q2_motion_ids} summarizes the motion IDs used at each scale.

\subsection{Knee negative power penalty.}
\label{sec:power}
For each knee joint $j$ with torque $\tau_j$ and joint velocity $\dot{q}_j$,
we define the instantaneous mechanical power
\(
 P_j = \tau_j \dot{q}_j
\)
and penalize excessive \emph{negative} power (large braking torques) beyond a
deadband threshold. Concretely, for knee joints selected via the regex
{\small\verb|".*_knee_joint"|} we compute
\begin{equation}
  \tilde{P}_j = \max\!\bigl(0,\; - P_j - 150\bigr),
\end{equation}
where the deadband is $150$ (in the same units as power, e.g., W).
The normalized per-step penalty is
\begin{equation}
  c_{\text{knee}} = \sum_{j \in \mathcal{J}_\text{knee}}
    \left(\frac{\tilde{P}_j}{500}\right)^{\!2},
\end{equation}
where $500$ is the normalization constant $\texttt{power\_norm}$.
In the reward function this term is added with weight $w=-10$, i.e.,
\(
 r_{\text{knee}} = w \, c_{\text{knee}}
\),
so that large negative joint power at the knees is strongly discouraged.

\subsection{TensorRT-Accelerated Deployment}
\label{sec:hdc-tensorrt-deploy}

For deployment on the real G1 robot we run the model as an ONNX graph, executed by ONNX Runtime with TensorRT acceleration. At run time the deploy node continuously receives joint states and IMU signals from the robot, builds a compact observation vector, and feeds it to the ONNX base policy.
The base policy outputs a 29-dimensional whole-body action, and, the residual policy produces a small corrective action in the same space; the two are added to obtain the final command.
This command is interpreted as desired joint positions, which are tracked by a PD controller with safety envelopes on torques and velocities, and converted into low-level motor commands at a control period of $20\,\mathrm{ms}$.

When a GPU with TensorRT is available, ONNX Runtime delegates the forward pass of these ONNX models to TensorRT, giving a lower-latency implementation of the same controller. In our implementation, we use \textbf{5 ODE integration steps}
(often informally referred to as denoising steps),
which strike a favorable balance between action quality and computational cost.
All inference is executed on the onboard
\textbf{NVIDIA Jetson Orin NX} of the Unitree G1 humanoid robot. With TensorRT acceleration, the end-to-end policy inference latency,
including the full ODE integration procedure,
is approximately \textbf{10\,ms} per control step. This latency comfortably satisfies real-time control requirements
and enables closed-loop execution of high-dynamic whole-body motions
entirely on the onboard compute, without reliance on offboard processing.

\subsection{Failure Cases and Discussion}
\label{sec:supp_failure}

While \model achieves strong sim-to-real performance overall, we observe a small
number of failure cases during real-world deployment.
These failures predominantly occur during highly impulsive landing phases of
certain extreme motions, where large transient braking loads trigger hardware
protection mechanisms, including motor overcurrent, power limits, or battery
undervoltage events.

Notably, these motions can be executed reliably in simulation and sim-to-sim
transfer, indicating that failures are not caused by degraded tracking accuracy
or loss of balance.
Instead, they expose residual discrepancies between simulated actuation models
and the true hardware capability envelope under extreme dynamic conditions. These observations suggest that further improving robustness for extreme motions
will require more comprehensive modeling of real actuator and power-system limits,
including the coupled effects of torque, speed, current, power flow, and battery
voltage dynamics, which remain challenging to capture accurately in simulation.

In the current work, post-training is performed via a lightweight residual policy that refines a frozen flow-based base controller. While this design offers stability and sample efficiency, it may also limit the extent to which the full representational capacity of the large flow-based model can be further adapted to hardware-specific constraints. Future research may explore more native post-training strategies~\cite{yi2026flow} that directly fine-tune or adapt the full base policy under actuation-aware objectives. Such approaches could potentially unlock greater expressivity and hardware alignment from large generative control models, enabling more principled integration between flow-based policy learning and real-world physical constraints.
\end{document}